\documentclass{article}
\pdfoutput=1
\PassOptionsToPackage{numbers, compress}{natbib}
\usepackage{iclr2020_conference,times}

\usepackage[utf8]{inputenc} % allow utf-8 input
\usepackage[T1]{fontenc}    % use 8-bit T1 fonts
\usepackage{hyperref}       % hyperlinks
\usepackage{url}            % simple URL typesetting
\usepackage{booktabs}       % professional-quality tables
\usepackage{amsfonts}       % blackboard math symbols
\usepackage{nicefrac}       % compact symbols for 1/2, etc.
\usepackage{microtype}      % microtypography

\usepackage{caption}
\usepackage{subcaption}
\usepackage{chngcntr}

\usepackage{xr}
\usepackage{amssymb}
\usepackage{mathtools}
\usepackage{multirow}
\usepackage{makecell}

\newcommand{\EE}{\mathbb{E}}

\newcommand{\AF}{\mathrm{AF}}
\newcommand{\F}{\mathrm{F}}
\newcommand{\EWC}{\mathrm{EWC}}

\newcommand{\SOS}{\mathrm{SOS}}

\newcommand{\MAS}{\mathrm{MAS}}
\newcommand{\SI}{\mathrm{SI}}
\newcommand{\SIU}{\mathrm{SIU}}

\newcommand{\SIB}{\mathrm{SIB}}

\newcommand{\appropto}{\mathrel{\vcenter{
			\offinterlineskip\halign{\hfil$##$\cr
				\propto\cr\noalign{\kern2pt}\sim\cr\noalign{\kern-2pt}}}}}

\title{Unifying Regularisation Methods for Continual Learning}

\author{Frederik Benzing\\
	Department of Computer Science\\
	ETH Zurich\\
	8092 Zurich, Switzerland \\
	\texttt{benzingf@inf.ethz.ch} \\
}

\begin{document}

	\maketitle
	%\addtocontents{toc}{\setcounter{tocdepth}{-10}}
	\begin{abstract}%   <- trailing '%' for backward compatibility of .sty file
Continual Learning addresses the challenge of learning a number of different tasks sequentially. The goal of maintaining knowledge of earlier tasks without re-accessing them starkly conflicts with standard SGD training for artificial neural networks. An influential method to tackle this problem without storing old data are so-called regularisation approaches. They measure the importance of each parameter for solving a given task and subsequently protect important parameters from large changes. In the literature, three ways to measure parameter importance have been put forward and they have inspired a large body of follow-up work. Here, we present strong theoretical and empirical evidence that these three methods, Elastic Weight Consolidation (EWC), Synaptic Intelligence (SI) and Memory Aware Synapses (MAS), are surprisingly similar and are all linked to the same theoretical quantity. Concretely, we show that, despite stemming from very different motivations, both SI and MAS approximate the square root of the Fisher Information, with the Fisher being the theoretically justified basis of EWC. Moreover, we show that for SI the relation to the Fisher -- and in fact its performance -- is due to a previously unknown bias. On top of uncovering unknown similarities and unifying regularisation approaches, we also demonstrate that our insights enable practical performance improvements for large batch training.
\end{abstract}

\section{Introduction}
\label{introduction}

Despite considerable progress, many gaps between biological and machine intelligence remain. Animals, for example, flexibly learn new tasks throughout their lives, while at the same time maintaining robust memories of their previous knowledge. This ability conflicts with traditional training procedures for artificial neural networks, which overwrite previous skills when optimizing new tasks \citep{mccloskey1989, goodfellow2013}. The field of continual learning is dedicated to mitigate this crucial shortcoming of machine learning. It exposes neural networks to a sequence of distinct tasks. When training for new tasks, the algorithm is not allowed to revisit old data. Nevertheless, it should retain previous skills and at the same time remaining flexible enough to incorporate new knowledge into the network.

An influential line of work to approach this challenge are so called regularisation-based algorithms. They remain to be the main approach to continual learning without expanding the network or storing old data, which is a key challenge and a main motivation of the field \citep{farquhar2018}.
The first regularisation-based method was introduced by \citet{kirkpatrick2017}, who proposed the Elastic Weight Consolidation (EWC) algorithm.
After training a given task, EWC measures the importance of each parameter for this task and introduces an auxiliary loss penalising large changes in important parameters. Naturally, this raises the question of how to measure `importance'. While EWC uses the diagonal Fisher Information, two main alternatives have been proposed: Synaptic Intelligence (SI, \citet{zenke2017}) aims to attribute the decrease in loss during training to individual parameters  and Memory Aware Synapses (MAS, \citet{aljundi2018}) introduces a heuristic measure of output sensitivity. Together, these three approaches have inspired many further regularisation-based approaches, including combinations of them \citep{chaudhry2018}, refinements \citep{huszar2018, ritter2018, chaudhry2018, yin2020sola}, extensions \citep{schwarz2018, liu2018rotate, park2019continual, lee2020continual} and applications in different continual learning settings \citep{aljundi2019task} as well as different domains of machine learning \citep{lan2019}. Almost every new continual learning method compares to at least one of the algorithms EWC, SI and MAS.

Despite their popularity and influence, basic practical and theoretical questions regarding these algorithms had previously been unanswered. Notably, it was unkown how similar these importance measures are. Additionally, for SI and MAS as well as their follow-ups there was no solid theoretical understanding of their effectiveness. Here, we close both these gaps through a  theoretical analysis confirmed by a series of carefully designed experiments on standard continual learning benchmarks. Our main findings can be summarised as follows:
\begin{enumerate}
	\item[(a)]\label{b} We show that SI's importance approximation is biased and that the bias rather than SI's original motivation is responsible for its performance. Further, also due to the bias, SI is approximately equal to the square root of the Fisher Information.
	\item[(b)]\label{a} We show that MAS, like SI, is approximately equal to the square root of the Fisher Information. 
	%\item{(c)} The previous two points
	\item[(c)] Together, (a) and (b) unify the three main regularisation approaches and their follow ups by explicitly linking all of them to the same theoretically justified quantity -- the Fisher Information.
	For  SI- and MAS-based algorithms this has the additional benefit of giving a more plausible theoretical explanation for their effectiveness.
	\item[(d)] Based on our precise understanding of SI, we propose an improved algorithm, Second-Order Synapses (SOS). We demonstrate that SOS outperforms SI in large batch training and discuss further advantages.
\end{enumerate}

	\section{Related Work}\label{sec:related_work}
The problem of catastrophic forgetting in neural networks has been studied for many decades \citep{mccloskey1989, ratcliff1990, french1999}.
In the context of deep learning, it received more attention again
\citep{goodfellow2013, srivastava2013}.
All previous work we are aware of proposes new or modified algorithms to tackle continual learning. No attention has been directed towards understanding existing methods and we hope that our work will not remain the only effort of this kind. %With the myriad of existing algorithms, we believe this to be a worthwhile goal. Below we survey some of the existing algorithms.

We now review the broad body of continual learning algorithms.
Following \citep{parisi2019}, they are  often categorised into regularisation-, replay- and architectural approaches. 

\textit{Regularisation methods} will be reviewed more closely in the next section. 

\textit{Replay methods} refer to algorithms which either store a small sample or generate data of old distributions and use this data while training on new methods \citep{rebuffi2017, lopez2017, shin2017, kemker2017}. These approaches can be seen as investigating how far standard i.i.d.-training can be relaxed towards the (highly non-i.i.d.)\ continual learning setting without losing too much performance. They are interesting, but usually circumvent the original motivation of continual learning to maintain knowledge \textit{without} accessing old distributions. Intriguingly, the most effective way to use old data appears to be simply replaying it, i.e.\ mimicking training with i.i.d.\ batches sampled from all tasks simultaneously \citep{chaudhry2019continual}.

\textit{Architectural methods} extend the network as new tasks arrive \citep{fernando2017, li2019, schwarz2018, golkar2019continual, von2019continual}. This can be seen as a study of how old parts of the network can be effectively used to solve new tasks and touches upon transfer learning. Typically, it avoids the challenge of integrating new knowledge into an existing networks.
Finally, \cite{van2019, hsu2018, farquhar2018} point out that  different continual learning scenarios and assumptions with varying difficulty were used across the literature.%They also clearly define and distinguish these scenarios and test how well different algorithms perform in them.
\footnote{In the appendix, we critically review and question some of their experimental results.}

	\section{Review of Regularisation Methods}\label{sec:desc}
%Here, we review the main regularisation approaches. %Related follow ups as well as alternative approaches to continual learning are reviewed in Appendix \ref{sec:related_work}. 

\textbf{Formal Description of Continual Learning.}
In continual learning we are given $K$ datasets $\mathcal{D}_1,\ldots,\mathcal{D}_K$ sequentially. When training a neural net with $N$ parameters $\theta\in \mathbb{R}^N$ on dataset $\mathcal{D}_k$, we have no access to the previously seen datasets $\mathcal{D}_{1:k-1}$. 
However, at test time the algorithms is tested on all $K$ tasks and the average accuracy is taken as measure of the performance.

\textbf{Common Framework for Regularisation Methods.}
Regularisation based  approaches introduced in  \cite{kirkpatrick2017, huszar2018} protect previous memories by modifying the loss function $L_k$ related to dataset $\mathcal{D}_k$. Let us denote the parameters obtained after finishing training on task $k$ by $\theta^{(k)}$ and let  $\omega^{(k)}\in \mathbb{R}^N$ be the parameters' \textit{importances}. When training on task $k$, regularisation methods use the loss 
\[
\tilde{L}_k = L_k + c\cdot \sum_{i=1}^N \omega_i ^{(k-1)}\left(\theta_i - \theta^{(k-1)}_i\right)^2
\]
where $c>0$ is a hyperparameter. The first term $L_k$ is the standard (e.g.\ cross entropy) loss of task $k$. The second term ensures that the parameters do not move away too far from their previous values.
Usually, $\omega_i^{(k)} = \omega_i^{(k-1)} + \omega_i$, where $\omega_i$ is the importance for the most recently learned task $k$.

\textbf{Elastic Weight Consolidation} \citep{kirkpatrick2017} uses the diagonal of the Fisher Information as importance measure. It is evaluated after training for a given task. To define the \textit{Fisher Information Matrix} $F$, we consider a dataset $\mathcal{X}$. 
For each datapoint $X\in \mathcal{X}$ the network predicts a probability distribution $\mathbf{q}_X$ over the set of labels $L$, where we suppress dependence of $\textbf{q}_X$ on $\theta$ for simplicity of notation. We denote the predicted probability of class $y$ by $\textbf{q}_X(y)$ for each $y\in L$. 
Let us assume that we minimize the negative log-likelihood $-\log \mathbf{q}_X(y)$ and write $g(X,y)=-\frac{\partial \log\textbf{q}_X(y)}{\partial \theta}$ for its gradient. 
Then the Fisher Information $F$ is given by
\begin{align}
\F &= 	 \mathbb{E}_{X\sim\mathcal{X}}	\mathbb{E}_ {y\sim \mathbf{q}_X} \left[g(X,y)g(X,y)^T
\right] \\
&= 
\mathbb{E}_{X\sim \mathcal{X}}	\sum_{y \in L} \textbf{q}_X(y) \cdot g(X,y)g(X,y)^T.
\label{fish}
\end{align}
Taking only the \textbf{diagonal Fisher} means 
$\omega_i (\EWC)= \mathbb{E}_{X\sim \mathcal{X}}\EE_{y\sim \textbf{q}_X} \left[g_i(X,y)^2\right].$
%The Fisher Information is often used as a measure of parameter sensitivity and  
Under the assumption that the learned label distribution $\textbf{q}_X$ is the real label distribution the Fisher equals the Hessian of the loss \citep{martens2014new, pascanu2013revisiting}. Its use for continual learning has a clear Bayesian, theoretical interpretation \citep{kirkpatrick2017, huszar2018}.

\textbf{Variational Continual Learning \citep{nguyen2017}} shares its Bayesian motivation with EWC, but uses principled variational inference in Bayesian neural networks. Its similarity to EWC and the fact that it uses `a smoothed version of the Fisher' is already discussed in \citep[][Appendix F]{nguyen2017} and further formalised in \citep{loo2020generalized}. We will therefore focus on the two algorithms below.

\textbf{Memory Aware Synapses} \citep{aljundi2018} heuristically argues that the sensitivity of the function output with respect of a parameter should be used as the parameter's importance. This sensitivity is evaluated after training a given task. 
The formulation in the paper suggests using the predicted probability distribution to measure parameter sensitivity. The probability distribution is the output of the neural network and the quantity that one aims to conserve during continual learning. Below we describe the precise importance this leads to. It was however brought to our attention that the MAS codebase seems to interpret the logits rather than the probability distribution as output of the neural network. In Appendix \ref{sec:MAS_logits} we show that the resulting version of MAS has the same performance as the one described here, and, crucially, is also related to the Fisher Information.\\
Denoting, as before, the final layer of learned probabilities by $\mathbf{q}_X$, the MAS importance is 
$$
\omega_i(\MAS) = \mathbb{E}_{X\sim \mathcal{X}}\left[  \left| \frac{\partial \|\textbf{q}_X\|^2}{\partial \theta_i} \right|\right],
$$

\textbf{Synaptic Intelligence} \citep{zenke2017} approximates the contribution of each parameter to the decrease in loss and uses this contribution as importance. To formalise the `contribution of a parameter', let us denote the parameters  at time $t$ by $\theta(t)$ and the loss by $L(t)$. If the parameters follow a smooth trajectory in parameter space, we can write the \textit{decrease} in loss from time $0$ to $T$ as
\begin{align}
L(0) - L(T) &=  - \int_{\theta(0)}^{\theta(T)} \frac{\partial L(t)}{\partial \theta} \theta'(t)  dt \\ 
&=  -\sum_{i=1}^N \int_{\theta_i(0)}^{\theta_i(T)} \frac{\partial L(t)}{\partial \theta_i} \theta_i'(t)  dt 
%\approx -\sum_{i=1}^N \sum_{t=0}^{T-1} \frac{\partial L(t)}{\partial \theta_i} \cdot \Delta_i(\theta)
.\label{def_contribution} 
\end{align}
%where the last approximation is a first order approximation of the integral. Note that the $i$-th summand in (\ref{def_contribution}) can be %interpreted as the contribution of parameter $\theta_i$ to the decrease in loss. Writing $\Delta_i(t)=(\theta_i(t+1)-\theta_i(t))$ we then get %the importance measure of SI:
The $i$-th summand in (\ref{def_contribution}) can be interpreted as the contribution of parameter $\theta_i$ to the decrease in loss.
While we cannot evaluate the integral exactly, we can use a first-order approximation to obtain the importances. 
To do so, we write $\Delta_i(t)=(\theta_i(t+1)-\theta_i(t))$ for an approximation of $\theta_i'(t) dt$ and get
\begin{eqnarray}\label{eq:def_SI}
\tilde{\omega}_i(\SI) =   \sum_{t=0}^{T-1} \frac{\partial L(t)}{\partial \theta_i} \cdot \Delta_i(t).
\end{eqnarray}
Thus, we have $\sum_i \tilde{\omega}_i(SI) \approx L(0) - L(T)$.
In addition, SI rescales its importances as follows\footnote{Note that the $\max(0,\cdot)$ is not part of the description in \citep{zenke2017}. However, we needed to include it to reproduce their results. A similar mechanism can be found in the official SI code.}
\begin{eqnarray}\label{eq:rescale}
\omega_i(\SI) = \frac{\max\left\{0, \tilde{\omega}_i(\SI)\right\}}{(\theta_i(T)-\theta_i(0))^2+\xi}.
\end{eqnarray}

\citet{zenke2017} use additional assumptions to justify this importance. One of the them -- using full batch gradient descent -- is violated in practice and we will show that this has an important consequence.

\textbf{Additional related regularisation approaches.} EWC, SI and MAS have inspired several follow ups.
We presented a version of EWC due to \citet{huszar2018} and tested in \citep{chaudhry2018, schwarz2018}. It is theoretically more sound and was shown to perform better. \citet{chaudhry2018} combine EWC with a `KL-rescaled'-SI.  \citet{ritter2018} use a block-diagonal (rather than diagonal) approximation of the Fisher; \citet{yin2020sola} use the full Hessian matrix for small networks. \cite{liu2018rotate} rotate the network to diagonalise the most recent Fisher Matrix, 
%allowing to take layer-wise parameter interactions into account. While improving performance, this comes at the cost of a memory requirement of order $N\cdot K$ (where N is the number of parameters and K the number of tasks) which is enough memory to train and store separate networks for each task. 
\citet{park2019continual} modify the loss of SI and \citet{lee2020continual} aim to account for batch-normalisation.

\section{Synaptic Intelligence, its Bias and the Fisher}
Here, we explain why SI is approximately equal to the square root of the Fisher, despite the apparent contrast between the latter and SI's path integral. 
First, we identify the bias of SI when approximating the path integral. We then show that the bias, rather than the path integral, explains similarity to the Fisher as well as performance. We carefully validate each assumption of our analysis empirically.

%\subsection{Theoretical Relation of SI and Absolute Fisher}
\subsection{Bias of Synaptic Intelligence}\label{bias}
To calculate $\omega(\SI)$ (\eqref{eq:def_SI}), we need to calculate the product $p = \frac{\partial L(t)}{\partial \theta}\cdot\Delta(t)$ for each $t$. Evaluating the full gradient $\frac{\partial L}{\partial \theta}$ is too expensive, so SI uses a stochastic minibatch gradient. The estimate of $p$ is biased since the same minibatch is used for the update $\Delta$ and the estimate of ${\partial L}/{\partial \theta}$. 

We now give the calculations detailing this argument.
For ease of exposition, let us assume vanilla SGD with learning rate $1$ is used. 
%Very similar arguments apply to other optimizers. 
Given a minibatch, denote its gradient estimate by $g+\sigma$, where $g=\partial L/{\partial \theta}$ denotes the full gradient and $\sigma$ the noise. The update is $\Delta =  g+\sigma$. Thus, $p$ should be  
$p = g \cdot(g+\sigma).$ 
However, using $g+\sigma$, which was used for the parameter update, to estimate $g$ results
 in 
$p_{bias} = (g+\sigma)^2.$ 
Thus, the gradient noise introduces a bias of
$\mathbb{E}[\sigma^2 +\sigma g] = \EE[\sigma^2].$

\paragraph{Unbiased SI (SIU).}
Having understood the bias, we can design an unbiased estimate by using two independent minbatches to calculate $\Delta$ and estimate $g$. We get $\Delta =  g+\sigma$ and an estimate $g+\sigma'$ for $g$ with independent noise $\sigma'$. We obtain
$
p_{no\_bias} = (g+\sigma') \cdot (g+\sigma) 
$
which in expectation equals $p = g \cdot(g+\sigma).$
Based on this we define an unbiased importance measure
$$
\tilde{\omega_i}(\SIU) =  \sum_{t=0}^{T-1} (g_t+\sigma_t') \cdot \Delta(t). 
$$
%where the dash indicated that the gradient estimate is independent of the parameter update.

\paragraph{Bias-Only version of SI (SIB).}\label{bias_only}
To isolate the bias, we can take the difference between biased and unbiased estimate. Concretely, this gives an importance which only measures the bias of SI %and is independent of the path integral, 
$$
\tilde{\omega_i}(\SIB) =   \sum_{t=0}^{T-1} ((g+\sigma)-(g+\sigma_t')) \cdot \Delta(t). 
$$ Observe that this estimate multiplies the parameter update $\Delta(t)$ with nothing but stochastic gradient noise. From the perspective of the SI path-integral, this should be meaningless and perform poorly. Our theory, detailed below, predicts differently.

\subsection{Relation of SI's Bias to Fisher}\label{sec:SI-OnAF}
The bias of SI depends on the optimizer used. The original SI-paper (and we) uses Adam \citep{kingma2014adam} and we now analyse the influence of this choice in detail; for other choices see Appendix \ref{sec:SOS_Advantage}. 
Recall that $\tilde{\omega}(SI)$ is a sum over terms $\frac{\partial{L(t)}}{\partial \theta} \cdot \Delta(t)$, where $\Delta(t) = \theta(t+1) - \theta(t)$ is the parameter update at time $t$. Both terms, $\frac{\partial{L(t)}}{\partial \theta}$ as well as $\Delta(t)$, are computed using the same minibatch. 
 Given a stochastic gradient $g_t+\sigma_t$, 
Adam keeps an exponential average of the gradient $m_t = (1-\beta_1)(g_t+\sigma_t) + \beta_1 m_{t-1}$ as well as  the squared gradient  $v_t = (1-\beta_2)(g_t+\sigma_t)^2 + \beta_2 v_{t-1}$. Ignoring minor normalisations, the parameter update is $\Delta(t) = \eta_t m_t/\sqrt{v_t}$, with learning rate $\eta_t=0.001,\beta_1=0.9$ and $\beta_2 = 0.999$. Thus, 
\begin{align}%\label{eq:noise}
	\frac{\partial L(t)}{\partial \theta} \Delta(t) &= \eta_t (1-\beta_1)
	%(g+\sigma)\cdot \frac{(1-\beta_1)(g+\sigma) + \beta_1 m_{t-1}}{\sqrt{v_t} + \epsilon} \\
	\frac{(g_t+\sigma_t)^2}{{\sqrt{v_t}}} + \eta_t \beta_1\frac{(g_t+\sigma_t)m_{t-1}}{{\sqrt{v_t}}} \nonumber\\
		&\stackrel{\textbf{(A1)}}{\approx} 
		\eta_t(1-\beta_1)\frac{(g_t+\sigma_t)^2}{{\sqrt{v_t}}} \label{eq:main_SI_bias}
\end{align}
Here, we made \textbf{Assumption (A1)} that the gradient noise is larger than the gradient, or more precisely: ${(1-\beta_1)\sigma_t^2\gg \beta_1 m_{t-1}g_t}$ (we ignore $\sigma_t m_{t-1}$ since $\EE[\sigma_t m_{t-1}]=0$ and since we average SI over many time steps). \textbf{(A1)} is equivalent to the bias of SI being larger than its unbiased part as detailed in Appendix \ref{sec:calc_bias}. Experiments in Section \ref{sec:SI-exp} and Appendix \ref{sec:noise} provide strong support for this assumption. Next, we rewrite
%This assumption is supported strongly by Figure \ref{figure:summed_MNIST}, where the difference between green (SI) and blue (SIU) line is caused precisely by $(1-\beta_1)\sigma^2$, see \ref{sec:calc_bias} for full details. We also refer to \citep{lan2019} and Figures \ref{figure:noise}, \ref{figure:noise_cdf} and \ref{figure:summed_cifar_all}, \ref{figure:summed_mnist_all}  for more evidence that the noise is considerably bigger than the gradient.
%are summed over time. 
%In addition, the denominator $\sqrt{v_t}$ changes only slowly (since $\beta_2=0.999$) 
%and is roughly equal to $\sqrt{\EE[(g_t+\sigma_t)^2]}$ across time. Therefore, 

\begin{align}\label{eq:SI-OnAF}
\tilde{\omega}(\SI) 
&\stackrel{\textbf{(A1)}}{\approx}
(1-\beta_1) \sum_{t\le T} \frac{\eta_t (g_t+\sigma_t)^2}{\sqrt{v_t}} \\
&= \frac{1-\beta_1}{\sqrt{v_T}} \sum_{t\le T} \eta_t\sqrt{\frac{v_T}{v_t}} (g_t+\sigma_t)^2
\end{align}

Now consider $v_t$. It is a decaying average of $(g_t+\sigma_t)^2$. When stochastic gradients become smaller during learning, $v_t$ will decay so that $\sum\sqrt{v_T/v_t}\cdot (g+\sigma_t)^2$ will be a (unnormalised, not necessarily exponentially) decaying average of the squared gradient. Therefore, we make \textbf{Hypothesis (A2)} 
\[
\sum_{t\le T}\sqrt{\frac{v_T}{v_t}}\cdot (g+\sigma_t)^2 \stackrel{\textbf{(A2)}}{\appropto} v_T,
\]
since both LHS \& RHS are decaying averages of the squared gradient. We will validate \textbf{(A2)} in Section \ref{sec:SI-exp}. Altogether, we obtain

\begin{align}\label{eq:SI-OnAF}
\tilde{\omega}(\SI) 
&\approx
\frac{1}{\sqrt{v_T}} \sum_{t\le T} \eta_t\sqrt{\frac{v_T}{v_t}} (g_t+\sigma_t)^2 \\
&\stackrel{\textbf{(A2)}}{\appropto}
\frac{v_T}{\sqrt{v_T}} = \sqrt{v_T}. \nonumber
\end{align}

To avoid confusion, we note that it may seem at first that a similar argument implies that SI is proportional to $v_t$ rather than $\sqrt{v_t}$. We explain in Appendix \ref{sec:SI_SOS-better} why this is not the case.

\begin{table*}[t]
	%\vskip -12pt
	\caption{\textbf{Test Accuracies on Permuted-MNIST and Split CIFAR} 
		%Algorithms are grouped by experiment as indicated in main text.
		(Mean and std-err of average accuracy over 3 resp.\ 10 runs for MNIST resp.\ CIFAR). Experiments No.\  (1) \& (3) explain SI's and MAS' performance. Exp (2) shows predicted improvements of our SOS for large batch training.
	}
	\label{table:results}
	%\vskip -0.6in
	\begin{center}
		%\vskip -0.15in
		\begin{small}
			%\vskip -0.1in
			%\begin{sc}
			%\vskip -0.1in
			\begin{tabular}{cccc|c|cccc}
				\toprule
				\textbf{No.} & \textbf{Algo.} & \textbf{P-MNIST} & \textbf{CIFAR} & &\textbf{No.} & \textbf{Algo.} & \textbf{P-MNIST} & \textbf{CIFAR} \\
				\midrule
				\multirow{4}*{(1)}
				& $\SI$ & 97.2$\pm$0.1& 74.4$\pm$0.2 &  
				&  \multirow{4}*{(3)} & $\MAS$ & 97.3$\pm$0.1& 73.7$\pm$0.2 
				\\
				& $\SI$ Bias-Only & 97.2$\pm$0.1& 75.1$\pm$0.1& 
				&& $\AF$ & 97.4$\pm$0.1& 73.4$\pm$0.1
				\\ 
				& $\SI$ Unbiased & 96.3$\pm$0.1& 72.5$\pm$0.3&
				&& $\sqrt{\text{Fisher}}$ & 97.1$\pm$0.2& 73.5$\pm$0.2
				\\ 
				& $\SOS$ & 97.3$\pm$0.1& 74.1$\pm$0.2&
				&& Fisher ($\EWC$) & 97.1$\pm$0.2 & 73.1$\pm$0.2 
				\\
				\midrule
				\multirow{2}*{(2)}
				& $\SI (2048)$ & 96.2$\pm$0.1& 70.0$\pm$0.3
				&&  &&&\multirow{2}*{} 
				\\
				& $\textbf{SOS}\mathbf{(2048)}$ & $\mathbf{97.1\pm0.1}$ & $\mathbf{74.4\pm0.1}$& 
				&&&&
				\\ 
				\bottomrule
			\end{tabular}
			%\end{sc}
		\end{small}
	\end{center}
\end{table*}

\textbf{Relation of SI to Fisher (EWC).}
Note that $v_t$ is a common approximation of the Hessian, e.g.\ \citep{graves2011practical, khan2018fast, aitchison2018bayesian},  and recall that the Hessian is approximately equal to the Fisher (see Appendix \ref{sec:SOS_calc} for a more detailed discussion).
Thus, other than using different approximations of the Fisher, the only difference between EWC and SI is the square root of the latter.\\
Even though we are not aware of a theoretical justification of the square root in this setting, this result explicitly links the bias and importance measure of SI to second order information of the Hessian. On a related note, we show in Appendix \ref{sec:sqrtF} that in a different regime the square root of the Hessian can be a theoretically justified importance measure.

\textbf{Influence of Regularisation.}
Note that gradients, second moment estimates and the update direction $\Delta (t)$ of Adam will be influenced by the auxiliary regularisation loss. In contrast, the approximation of $\partial L /\partial \theta$ in \eqref{eq:main_SI_bias} only relies on current task gradients (without regularisation). 
Thus, the relation between SI and $\sqrt{v_T}$ (which in this context is decaying average of task-only gradients), will be more noisy in the presence of large regularisation. 
%We take $v_t$ to be the running
%The derivation above assumes that the parameter update $\Delta(t)$ is given by the gradients of the current task, ignoring the auxiliary regularisation loss. The latter will change update direction and momentum of Adam. In contrast, when we calculate ${v_T}$  as importance measure (not as part of the optimizer), we use only gradients of the current task. Thus, strong regularisation will make the relation between $\omega(\SI)$ and $\sqrt{v_t}$ noisy, but not abolish it. 
This will be confirmed empirically in Section \ref{sec:SI-exp}.
%See Fig \ref{figure:correlations} (middle row) and Appendix \ref{sec:effect_reg_SI} for various confirmations hereof.% for controls. %experiments.%showing that SI, SIB are more related to AF than SIU (and more effective at continual learning).

\textbf{Practical Implications.}  A benefit of our derivation is that it makes the dependence of SI on the optimisation process explicit. Consider for example the influence of batch size. A priori one would not expect it to affect SI more than  other methods. Given our derivation, however, one can see that a larger batch size reduces noise and thus bias, making \textbf{(A1)} less valid. Moreover, $v_t$ will be a worse approximation of the Fisher (see Appendix \ref{sec:SOS_calc} or e.g.\ \citep{khan2018fast}). Thus, if SI relies on its relation to second order information of the Fisher, then we expect it to suffer disproportionately from large batch sizes -- a prediction we will confirm empirically below.
Additionally, learning rate decay, choice of optimizer and training set size and difficulty could harm SI as discussed in Appendix \ref{sec:SOS_Advantage}.

\textbf{Improving SI: Second-Order Synapses (SOS).} Given our derivation, we propose to adapt SI to explicitly measure second order information contained in the Fisher. To this end, we propose the algorithm Second-Order Synapses (SOS). In its simplest form it uses $\omega(\SOS)=\sqrt{v_t}$ as importance. For larger batch sizes we introduce a similar, but provably better approximation of the Fisher detailed in Appendix \ref{sec:SOS_calc}. We will confirm some benefits of SOS over SI empirically below and discuss additional advantages in Appendix \ref{sec:SOS_Advantage}. Note also that SOS is computationally more efficient than MAS and EWC as it does not need a pass through the data after training a given task.

%MAS, \EE[|g|], sqrt{Fisher}, Fisher 
%SI, SIB, SIU, SOS
%SI(2048), SOS(2048)

\subsection{Empirical Investigation of SI, its bias and Fisher}\label{sec:SI-exp}
%Our first important insight was recognising the bias of SI, which we now investigate empirically.

\begin{figure*}[h]
	%\vskip -8pt
	\begin{subfigure}{0.3\textwidth}
		\begin{center}
			\centerline{\includegraphics[width=\columnwidth]{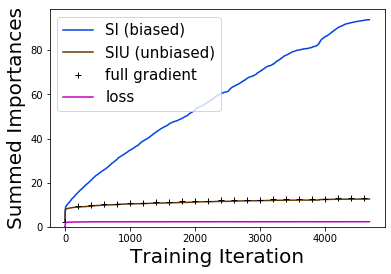}}
		\end{center}
	\end{subfigure}
	\hskip 12pt
	\begin{subfigure}{0.3\textwidth}
		\begin{center}
			\centerline{\includegraphics[width=\columnwidth]{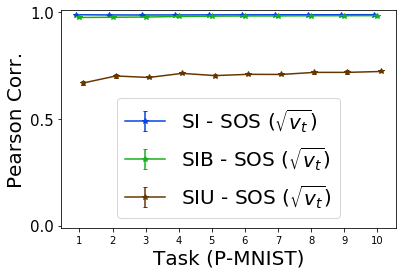}}
		\end{center}
	\end{subfigure}
	\hskip 12pt
	\begin{subfigure}{0.3\textwidth}
		\begin{center}
			\centerline{\includegraphics[width=\columnwidth]{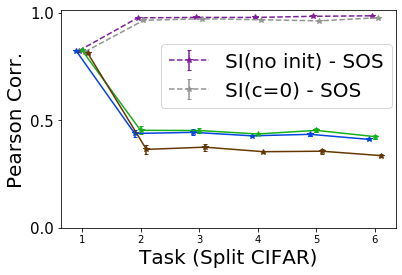}}
		\end{center}
	\end{subfigure}

	\begin{subfigure}{0.3\textwidth}
	\begin{center}
		\centerline{\includegraphics[width=\columnwidth]{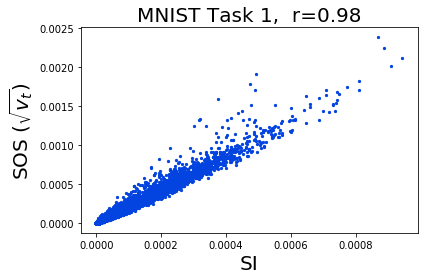}}
	\end{center}
\end{subfigure}
\hskip 12pt
\begin{subfigure}{0.3\textwidth}
	\begin{center}
		\centerline{\includegraphics[width=\columnwidth]{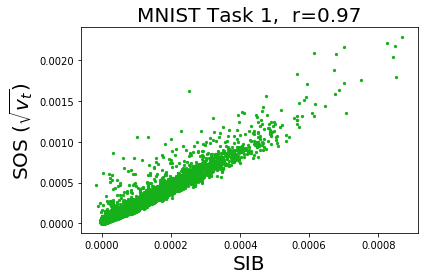}}
	\end{center}
\end{subfigure}
\hskip 12pt
\begin{subfigure}{0.3\textwidth}
	\begin{center}
		\centerline{\includegraphics[width=\columnwidth]{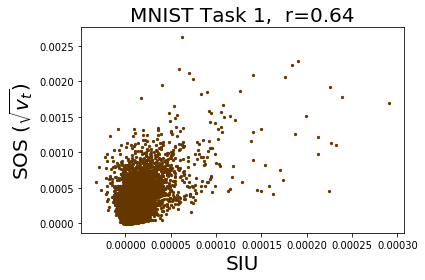}}
	\end{center}
\end{subfigure}

	%\vskip -0.2in
	\caption{\textbf{SI, Bias and Square Root of Approximate Fisher $v_t$ (SOS).}\\
		\textbf{Top Left:} Summed Importances for SI and its unbiased version, showing that the bias dominates SI and that Assumption \textbf{(A1)} holds. \\
		\textbf{Top Center:} Pearson correlations of SI, its bias (SIB), and unbiased version (SIU) with SOS, showing that relation between SI and SOS is strong and due to bias; confirming \textbf{(A2)}. The same is shown by the scatter plots.\\
		\textbf{Top Right:} Same as Mid but on CIFAR; additionally, relation between SOS and two SI-controls is shown: `no init'  does not re-initialise network weights after each task; `$c=0$' has regularisation strength $0$. This shows that strong regularisation weakens the tie between SI and SOS on Tasks 2-6 as explained by our theory. \\
		\textbf{Bottom:} Scatter plots of SOS ($\sqrt{v_t}$) with SI (left), its bias-only SIB (middle) and its unbiased version SIU (right); showing $10^5$ randomly sampled weights. A straight line through the origin corresponds to two importance measures being multiples of each other as was suggested by our derivation for SI and SOS, but not for SIU and SOS. Note that SIU has negative importances before rescaling in \eqref{eq:rescale}.
	}
	\label{figure:SI}
	%\vskip -0.1in
\end{figure*}

\textbf{Bias dominates SI.} According to the motivation of SI the sum of importances over parameters $\sum_i \tilde{\omega}_i(\SI)$ should track the decrease in loss $L(0)-L(T)$, see \eqref{def_contribution}. Therefore, we investigated how well the summed importances of SI and its unbiased version SIU approximate the decrease in loss. We also include an approximation of the path integral which uses the full training set gradient $\partial L / \partial \theta$.
The results in Figure \ref{figure:SI} (left) show:  
 (1) Using an unbiased gradient estimate and the full gradient gives almost identical sums, supporting the validity of the unbiased estimator.\footnote{Note that even the unbiased first order approximation of the path integral overestimates the decrease in loss. This is consistent with findings that the loss has positive curvature \citep{jastrzebski2018relation, lan2019}.}
 (2) The bias of SI is 5-6 times larger than its unbiased so that SIU yields a considerably better approximation of the path integral 
 \\  
%Note that the large magnitude of the bias (i.e.\ the noise) also shows that our approximation in eq \eqref{eq:noise} is valid, see \ref{sec:calc_bias} for full details. 

\textbf{Checking (A1).}
We point out that the bias, i.e.\ the difference between SI and SIU, in Figure \ref{figure:SI} (left) is due to the term
$(1-\beta_1)\sigma_t^2$. Thus, the fact that the bias is considerably larger then the unbiased part is direct, strong evidence of Assumption \textbf{(A1)} (see Appendix \ref{sec:calc_bias} for full calculation). 
Results on CIFAR are analogous (Figure \ref{figure:summed_cifar_all} ).

\textbf{Checking (A2): SI is almost equal to the Square Root of Approximate Fisher $v_t$.} To check \textbf{(A2)}, we compared $\omega(\SI)$ and $\omega(\SOS) = \sqrt{v_t}$, see Figure~\ref{figure:SI} (mid, right). Correlations are almost equal to 1 on MNIST. 
The same holds on CIFAR Task 1, where the slight drop in correlation from around 0.99 to 0.9 is only due to the division in equation \eqref{eq:rescale} (c.f.\ Appendix \ref{moreisbetter}). In summary, 
this shows that  $\sqrt{v_t}$ is a very good approximation of SI. Correlations of SI to SOS on CIFAR Task 2-6 decrease due to regularisation, see below.

\textbf{Effect of Regularisation.}
The drop of correlations on CIFAR tasks 2-6 is due to large regularisation as explained theoretically in Section \ref{sec:SI-OnAF} and confirmed by two controls of SI with less strong regularisation:
The first control simply sets regularisation strength to $c=0$. The second control refrains from re-initialising the network weights after each task (exactly as in original SI, albeit with slightly worse validation performance than the version with re-initialisation). In the second setting the current parameters $\theta$ never move too far from their old value $\theta^{(k-1)}$, implying smaller gradients from the quadratic regularisation loss, and also meaning that a smaller value of $c=0.5$ is optimal. We see that for both controls with weaker regularisation the correlation of SI to SOS is again close to one, supporting our theory. 

\textbf{Bias explains SI's performance.} We saw that SI's bias is much larger than its unbiased part. But how does it influence SI's performance? To quantify this, we compared SI to its unbiased version SIU and the bias-only version SIB. Note that SIB is completely independent of the path integral motivating SI, only measures gradient noise and therefore should perform poorly according to SI's original motivation. However, the results in Table \ref{table:results}(1), reveal the opposite: Removing the bias reduces performance of SI (SIU is worse), whereas isolating the bias does not affect or slightly improve performance. 
%(difference between SIB and SI on CIFAR is statistically significant, $p<0.005$ two-sided t-test). 
This demonstrates that SI relies on its bias, and not on the path integral, for its continual learning performance.

\textbf{Bias explains Second-Order Information.} We have seen that the bias, not the path integral, is responsible for SI's performance. Our theory offers an explanation for this surprising finding, namely that the bias is responsible for SI's relation to the Fisher. But does this explanation hold up in practice? To check this, we compare SI, its unbiased variant SIU and its bias-only variant SIB to the second order information $\omega(\SOS) = \sqrt{v_T}$. The results in Figure~\ref{figure:SI} (top middle\&right; bottom) show that SI, SIB are more similar to $\sqrt{v_T}$ than SIU, directly supporting our theoretical explanation. 

Together the two previous paragraphs are very strong evidence that SI relies on the second order information contained in its bias rather than on the path integral. 

\textbf{SOS improves SI.} We predicted that SI would suffer disproportionately from training with large batch sizes. To test this prediction, we ran SI and SOS with larger batch sizes of 2048, Table \ref{table:results} (2). We found that, indeed, SI's performance degrades by roughly $1\%$ on MNIST and more drastically by $4\%$ on CIFAR. In contrast, SOS' performance remains stable.\footnote{We emphasise that the difference of SI and SOS is \textit{not} explained by a change in `training regime' \citep{mirzadeh2020understanding}, since both algorithms use equally large minibatches.} This does not only validate our prediction but demonstrates that our improved understanding of SI leads to notable performance improvements. See Appendix \ref{sec:SOS_calc} for a discussion of additional advantages of SOS.

\begin{figure*}[t]
	%\vskip -40pt
	\begin{center}
		\begin{subfigure}[b]{0.3\textwidth}
			\centerline{\includegraphics[width=\textwidth]{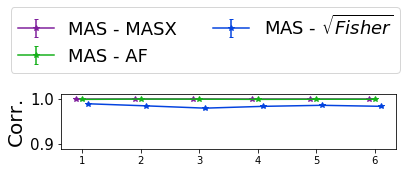}}
			%\vskip 1in
			\centerline{\includegraphics[width=\textwidth]{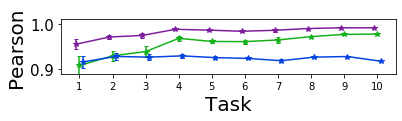}}
		\end{subfigure}
		\hskip 12pt
		\begin{subfigure}[b]{0.3\textwidth}
			\centerline{\includegraphics[width=\textwidth]{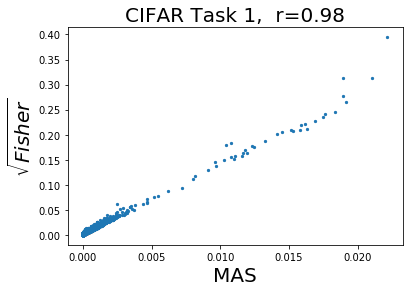}}
		\end{subfigure}
		\hskip 12pt
		\begin{subfigure}[b]{0.3\textwidth}
			\centerline{\includegraphics[width=\textwidth]{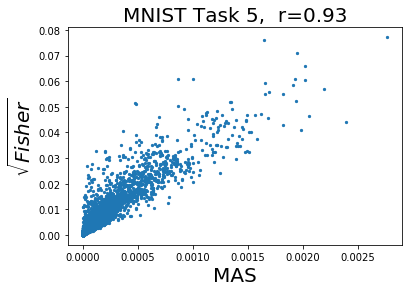}}
		\end{subfigure}
	\end{center}
	%\vskip -16pt
	\caption{
		\textbf{Empirical Relation between MAS and Square Root of Fisher.} \\
		\textbf{Left}: Summary of Pearson Correlations (top: CIFAR, bottom: MNIST), supporting Assumptions \textbf{(B1)-(B3))}.\\ \textbf{Mid~\&~Right}: Scatter plots of importance measures. Each point in the scatter plot corresponds to one weight of the net, showing $10^5$ randomly sampled weights. A straight line through the origin corresponds to two importance measures being multiples of each other as was suggested by our theoretical analysis.
	}
	\label{figure:MAS}
	%\vskip -0.2in
\end{figure*}

\section{Memory Aware Synapses (MAS) and Fisher}
Here, we explain why -- despite its different motivation and like SI -- MAS is approximately equal to the square root of the Fisher Information. We do so by first relating MAS to $\AF:=\EE[|g+\sigma|]$ and then theoretically showing how this is related to the Fisher $F = \EE[|g+\sigma|^2]$ under additional assumptions.
As before, we check our theoretical derivations empirically.

\subsection{Theoretical Relation of MAS and Fisher}
We take a closer look at the definition of the importance of MAS. Recall that we use the predicted probability distribution of the network to measure sensitivity rather than logiits. With linearity of derivatives, the chain rule and writing $y_0 = \text{argmax} \mathbf{q}_X$,  we see (omitting expectation over $X\sim\mathcal{X}$)
\begin{align}\label{MAS1}
 \left|\frac{\partial \|  \textbf{q} \|^2}{\partial\theta}  \right| 
&= 2\left|\sum_{y\in Y} \textbf{q}(y) \frac{\partial \textbf{q}(y)}{\partial \theta}  \right| 
 \stackrel{\textbf{(B1)}}{\approx} 
2\left|\textbf{q}(y_0)\frac{\partial \textbf{q}(y_0)}{\partial \theta} \right| \nonumber\\
&=2 \textbf{q}^2(y_0)\left|\frac{\partial \log \textbf{q}(y_0)}{\partial \theta}\right| %= 2 \textbf{q}^2_X(y_0) g(X,y_0)
\end{align}
Here, we made \textbf{Assumption (B1)} that the sum is dominated by its maximum-likelihood label $y_0$, which should be the case if the network classifies its images confidently, i.e.\ $\textbf{q}(y_0)\gg \textbf{q}(y)$ for $y\neq y_0$. Recall that the importance is measured at the end of training, so that this assumption is justified  if the task has been learned successfully.
Using the same assumption (B1) we get %(akin to the \textit{Predictive Fisher})
\begin{align}
\EE_{y\sim \textbf{q}_X}\left[|g+\sigma|\right] 
\stackrel{\textbf{(B1)}}{\approx}
 \textbf{q}(	y_0) |g+\sigma| = \textbf{q}(y_0) \left|\frac{\partial\log \textbf{q}(y_0)}{\partial\theta}\right|.\nonumber
\end{align}
showing that the only difference between $\omega(\MAS)$ and $\EE[|g|]$ is a factor of $2\textbf{q}_X(y_0)$. It is reasonable to make \textbf{Assumption (B2)} that this factor is approximately constant, since the model learned to classify training images confidently. This leads to \textbf{(B2)}:~$\omega(\MAS) \stackrel{}{\appropto} \AF = \EE[|g|]$. Note that
even with a pessimistic guess that $\textbf{q}_X(y_0)$ is in a range of 0.5 (rather inconfident) and 1.0 (absolutely confident), the two measures would be highly correlated.  

Now, it remains to explore how $\EE[|g+\sigma|]$ is related to the Fisher $\F = \EE[|g+\sigma|^2]$.  The precise relationship between the two will depend on the distribution of $g+\sigma$.
If, for example, gradients are distributed normally $(g+\sigma)\sim\mathcal{N}(\mu, \Sigma)$ with $\Sigma_{i,i} \gg \mu_i$ (corresponding to the observation that the noise is much bigger than the gradients, which we have seen to be true above), then we obtain $\F_i \approx \Sigma_{i,i}$ and, since $|g+\sigma|$ follows a folded normal distribution, we also have $\EE[|g|] \approx c \sqrt{\Sigma_{ii}}$ with $c=\sqrt{2/\pi}$. Thus, in the case of a normal distribution with large noise, we have $\AF=\EE[|g|] \appropto \sqrt{\F}$. Note that the same conclusion arises in under different assumptions, too, e.g.\ assuming that gradients follow a Laplace distribution with  $scale \gg mean$. 

\textbf{Relation of Fisher and MAS.} The above analysis leads to \textbf{Hypothesis~(B3)}:~$\omega(\MAS) \stackrel{}{\appropto} \sqrt{\text{Fisher}}$. %Eventually, this can only be evaluated empirically since precise knowledge of the gradient distribution is not attainable and we

\subsection{Empirical Relation of MAS and Fisher}
First, to assess \textbf{(B1)} in \eqref{MAS1} we compare LHS (MAS) and RHS (referred to as MASX) at the end of each task. We consistently find that the pearson correlations are almost equal to 1 strongly supporting \textbf{(B1)}, see Figure~\ref{figure:MAS}.

The correlations between $\omega(\MAS)$ and $\AF = \EE[|g+\sigma|]$ are similarly high, confirming \textbf{(B2)}.

The correlations between $\omega(\MAS)$ and the square root of the Fisher $\sqrt{\F}$ are around 0.9 for Permuted MNIST and approximately 1 for all CIFAR task. This confirms our theoretical hypothesis \textbf{(B3)} that MAS is approximately equal to the square root of the Fisher. 

In addition, to check whether similarity to the square root of the Fisher can serve as an explanation for performance of MAS, we ran continual learning algorithms based on MAS, AF and $\sqrt{\F}$. The fact that these algorithms perform similarly, Table~\ref{table:results}~(3), is in line with the claim that similarity to $\sqrt{\F}$ provides a theoretically plausible explanation for MAS' effectiveness.

	\section{Experimental Setup}\label{sec:exp_overview}
We outline the experimental setup, see Appendix \ref{sec:exp_detail} for details. 
It closely follows SI's setting \citep{zenke2017}: In domain-incremental \textbf{Permuted MNIST} \citep{goodfellow2013} each of 10 tasks consists of a random (but fixed) pixel-permutation of MNIST  and a \textbf{fully connected ReLU network} is used. The task-incremental \textbf{Split CIFAR 10/100} \citep{zenke2017} consists of six 10-way classification tasks, the first is CIFAR 10, and the other ones are extracted from CIFAR 100. The \textbf{keras default} CIFAR 10 \textbf{convolutional architecture} (with dropout and max-pooling) is used \cite{kreasdefault}.   
The only difference to \citet{zenke2017} is that like \citet{swaroop2019}, we usually re-initialise network weights after each task, observing better performance.
Code is available on github. %\footnote{\url{https://github.com/freedbee/continual_regularisation}}. 
%For brevity, we partly showed plots for only one benchmark, see appenidx for analogous plots. %Analogous plots for the other benchmark can then be found in the appendix.

	\section{Discussion}\label{sec:rel}
We have investigated regularisation approaches for continual learning, which are the method of choice for continual learning without replaying old data or expanding the model. We have provided strong theoretical and experimental evidence that both MAS and SI approximate the square root of the Fisher Information.
While the square root of the Fisher has no clear theoretical interpretation itself, our analysis makes explicit how MAS and SI are related to second-order information contained in the Fisher. This provides a more plausible explanation of the effectiveness for MAS- and SI based algorithms.
In addition, it shows how the three main regularisation methods are related to the same theoretically justified quantity, providing a unified view of these algorithms and their follow ups.

Moreover, our algorithm SIU to approximate SI's path integral can be used for the (non continual learning) algorithm LCA \citep{lan2019}, which relies on an expensive approximation of the same integral. This opens up new opportunities to apply LCA to larger models and datasets.

For SI, our analysis included uncovering its bias. We found that the bias explains performance better than the path integral motivating SI. Our theory offers a sound, empirically confirmed explanation for this otherwise surprising finding. 
Beyond this theoretical contribution, by proposing the algorithm SOS we gave a concrete example how understanding the inner workings of SI leads to substantial performance improvements for large batch training. We have dedicated Appendix \ref{sec:SOS_Advantage} to describing further potential improvements granted by SOS.

So far, studies providing a better understanding of regularisation algorithms and their similarities had been neglected. Our contribution fills this gap and, as exemplified above, 
proves to be a useful tool to predict, understand and improve these algorithms.

\section*{Acknowledgements}
I would like to thank Xun Zou, Simon Schug, Florian Meier, Angelika Steger, Kalina Petrova and Asier Mujika for helpful discussions and valuable feedback on this manuscript. I am also thankful for AM's technical advice and the Felix Weissenberger Foundation's ongoing support.

	\bibliography{./ic/my_refs}
	\bibliographystyle{iclr2020_conference}
	
	\newpage
	%\addtocontents{toc}{\setcounter{tocdepth}{4}} 
	\appendix
	\counterwithin{figure}{section}
	\counterwithin{table}{section}
	\tableofcontents
	\section{Overview over Algorithms}\label{sec:tabular_overview}
A tabluar overview of algorithms is given in Table \ref{table:overview}
\begin{table}[h]
	%\vskip -32pt
	\caption{\textbf{Summary of Regularisation Methods and Related Baselines.}\\ 
		\textit{Notation and Details:}
		Algorithms on the top calculate importance `online' along the parameter trajectory during training. Algorithms on the bottom calculate importance at the end of training a task by going through (part of) the training set again. Thus, the sum is over timesteps $t$ (top) or datapoints $X$ (bottom). $N$ is the number of images over which is summed. For a datapoint $X$, $\textbf{q}_X$ denotes the predicted label distribution and 
		$g(X,y)$ refers to the gradient of the negative log-likelihood of $(X,y)$. 
		$\Delta(t)= \theta(t+1) - \theta(t)$ refers to the parameter update at time $t$, which depends on both the current task's loss and the auxiliary regularisation loss. Moreover, 
		$(g_t+\sigma_t)$ refers to the stochastic gradient estimate of the current task's loss (where $g_t$ is the full gradient and $\sigma_t$ the noise) given to the optimizer to update parameters. In contrast, $(g_t + \sigma_t')$ refers to an independent stochastic gradient estimate. \\
		Algorithms SI, SIU, SIB rescale their final importances as in equation \eqref{eq:rescale} for fair comparison.\\
		For description, justification and choice of $\alpha$ in SOS, see Appendix \ref{sec:SOS_calc}.
	}\label{table:overview}
	\iffalse
	\caption{\textbf{Overview of different algorithms and their importance measures for one task.} 
		Algorithms on the top calculate importance `online' along the parameter trajectory during training. Algorithms on the right calculate importance at the end of training a task by going through (part of) the training set again: The sum is over timesteps $t$ (top) or datapoints $X$ (bottom). $N$ is the number of images over which is summed. Note that SI, SIU, SIB rescale their final importances as in \eqref{eq:rescale}.
		In all algorithms $\Delta(t)= \theta(t+1) - \theta(t)$ refers to the parameter update at time $t$, which depends on both the current task's loss and the regularisation loss. Moreover, 
		$(g_t+\sigma_t)$ refers to the stochastic gradient estimate of the current task's loss (where $g_t$ is the full gradient and $\sigma_t$ the noise) given to the optimizer (together with the regularisation loss) to calculate the parameter update. In contrast, $(g_t + \sigma_t')$ refers to an independent stochastic gradient estimate of the current task loss. For description, justification and choice of $\alpha$ in SOS, see Appendix \ref{x}.
		For a datapoint $X$, $\textbf{q}_X$ denotes the predicted label distribution and 
		$g(X,y)$ refers to the gradient of the negative log-likelihood of $(X,y)$. }
	\fi
	%\vskip -0.5in
	\centering
	\begin{small}
		\begin{tabular}{ll}
			\toprule
			\textbf{Name}  & \textbf{Paramater Importance } $\mathbf{\omega(\cdot)}$  
			\\ \midrule
			SI &  $\displaystyle\sum_{t}^{} (g_t+\sigma_t)\Delta(t)$ 
			\\
			%\midrule
			SIU (SI-Unbiased)  & $\displaystyle\sum_t (g_t+\sigma_t')\Delta(t)$ 
			\\
			%\midrule
			SIB (SI Bias-only)&  $\displaystyle\sum_t (\sigma_t- \sigma_t')\Delta(t)$ 
			\\
			%	\midrule					
			SOS (simple)  & $\displaystyle\frac{1-\beta_2}{1-\beta_2^{T+1}}\sum_{t\le T} \beta_2^{T-t}(g_t+\sigma_t)^2$ 
			\\
			SOS (2048, unbiased)  & $\frac{1-\beta_2}{1-\beta_2^{T+1}}\displaystyle\sum_{t\le T} \beta_2^{T-t}\bigl((g_t+\sigma_t) - \alpha(g_t+\sigma'_t) \bigr)^2$ 
			%&$\surd$ 
			\\ 
			\midrule 
			Fisher (EWC)  & $\displaystyle\frac1N\sum_X \EE_{y\sim \textbf{q}_X}\left[g(X,y)^2 \right] $ %&$\surd$ 
			\\
			%\midrule
			AF  & $\displaystyle\frac1N\sum_X \EE_{y\sim \textbf{q}_X}\left[|g(X,y)| \right] $
			%&$\surd$
			\\
			%\midrule
			MAS & $\displaystyle\frac1N\sum_X \left|\frac{\partial ||\textbf{q}_X||^2}{\partial \theta}\right| $
			%&$\surd$ 
			\\
			\bottomrule
		\end{tabular}
	\end{small}
	\vskip -0.15in
\end{table}
	
\section{Potential Advantages of SOS vs SI}\label{sec:SOS_Advantage}
Here, we discuss some problems that SI may have and for which SOS offers a remedy due to its more principled nature. We also explain how the points discussed here could explain findings of a large scale empirical comparison of regularisation methods \citep{de2019continual} and leave testing these hypotheses to future work. We note that it seems difficult/impossible to even generate hypothesis for some of the findings in \cite{de2019continual} without our contribution. Further, if these hypotheses are true, the described shortcomings are addressed by SOS.
\begin{enumerate}
	\item \textbf{Trainingset Size and Difficulty:} Recall that SI is similar to an unnormalised decaying average of $(g+\sigma)^2$. Since the average is unnormalised, the constant of proportionality between $\omega(\SI)$ and $\sqrt{v_t}$ will depend: (1) On the number of summands in the importance of SI, i.e.\ the number of training iterations and (2) On the speed of gradient decay, which may well depend on the training set difficulty. Thus, SI may underestimate the importance of small datasets (for which fewer updates are performed if the number of epochs is kept constant as e.g.\ in \cite{zenke2017, de2019continual}) and also for easy datasets with fast gradient decay. \\
	How would we expect this to affect performance?
	If easy datasets are presented first, then SI will undervalue their importance and thus forget them later on. If hard tasks are presented first, SI will overvalue their importance and make the network less plastic -- as latter tasks are easy, this is not too big of a problem. So if easy tasks are presented first, we expect SI to perform worse on average and forget more than when hard tasks are presented first. Strikingly, both lower average accuracy as well as higher forgetting are exactly what \citet[][Table 4] {de2019continual} observe for SI when easy tasks are presented first (but not for other methods). 
	\item \textbf{Learning Rate Decay:} In our derivation in Section \ref{sec:SI-OnAF} we assumed a constant learning rate $\eta_t = \eta_0$. If, however, learning rate decay is used, then this will counteract the importance of SI being a decaying average of $(g+\sigma)^2$; for a deacying leanring rate more weight will be put on earlier gradients, which will likely make the estimate of the Fisher less similar to the Fisher at the end of training. This may explain why SI sometimes has worse performance than EWC, MAS in experiments of \citet{de2019continual}
	\item \textbf{Choice of Optimizer:} Note that the bias and thus importance of SI depend on the optimizer. For example if we use SGD with learning rate $\eta_t$ (with or without momentum), following the same calculations as before shows that $\omega(\SI) \approx \sum_t \eta_t(g_t+\sigma_t)^2$. On the one hand, this is related to the Fisher so it may give good results, on the other hand it may overvalue gradients early in training, especially when combined with learning rate decay. See Appendix \ref{moreisbetter} for an experiment with SGD. 
	\citet{de2019continual} use SGD with learning rate decay, which again may be why SI sometimes performs worse than competitors.\\
	In an extreme case, when we approximate natural gradient descent by preconditioning with the squared gradient, the SI importance will be constant across parameters, showcasing another, probably undesired, dependence of SI on the choice of optimizer.
	\item \textbf{Effect of Regularisation:} We already discussed in the main part how strong regularisation influences SI's importance measure. In fact, on CIFAR tasks 2-6 many weights have negative importances, since the regularistation gradient points in the opposite direction of the task gradients and is larger, compare e.g.\ Figure \ref{figure:SI} and \ref{figure:correlations_SI_divide}. Negative importances seem counterproductive in any theoretical framework.
	\item \textbf{Batch Size:} We already predicted and confirmed that large batchsizes hurt SI and proposed a remedy to this issue, see also Appendix \ref{sec:SOS_calc}.
	
\end{enumerate}

\section{Second Moment of Gradients, Hessian, Fisher and SOS}\label{sec:SOS_calc}
As pointed out in the main paper, the second moment $v_t$ of the gradient is a common approximation of the Hessian. Here, we briefly review why this is the case, why the approximation becomes worse for large batch sizes and show how to get a better estimate. We note that the relations between Fisher, Hessian, and squared gradients, which are recapitulated here, are discussed in several places in the literature, e.g.\ \citep{martens2014new, kunstner2019limitations, thomas2020interplay}.

One way to relate the squared gradients to the Hessian is through the Fisher\footnote{Alternatively, one can directly apply the derivation, which is used to relate Fisher and Hessian, to the squared gradient ( Empirical Fisher).}: The Fisher Information is an approximation of the Hessian, which becomes exact when the learned label distribution coincides with the real label distribution. The Fisher takes an expectation over the model's label distribution. A common approximation is to replace this expectation by the (deterministic) labels of the dataset. This is called the \textit{Empirical Fisher} and it is a good approximation of the FIsher if the model classifies most (training) images correctly and confidently. 
To summarise, the Fisher is a good approximation of the Hessian under the assumption that the model's predicted distribution coincides with the real distribution and - under the same assumption - the empirical Fisher is a good approximation of the Hessian. In fact, taking a closer look at the derivations, it seems plausible that the empirical Fisher is a better approximation of the Hessian than the real Fisher, since - like the Hessian - it uses the real label distribution rather than the model's label distribution.

To avoid confusion and as pointed out by \cite{martens2014new}, we also note that the Empirical Fisher is not generally equal to the Generalised Gauss Newton matrix.
%Another way is through the Generalised Gauss-Newton Matrix. The diagonal of the Gauss Newton Matrix contains squared gradients, however in this case the logits (or the representation before the )

Note that the connection between squared gradients and the Hessian assumes that gradients are squares gradients before averaging. Here, and in many other places e.g.\ \citep{khan2018fast, aitchison2018bayesian}, this is approximated by squaring after averaging over a mini-batch since this is easier to implement and requires less computation. We describe below why this approximation is valid for small batch sizes and introduce an improved and easy to compute estimate for large batch sizes. We have not seen this improved estimate elsewhere in the literature.

For this subsection, let us slightly change notation and denote the images by $X_1,\ldots, X_D$ and the gradients (with respect to their labels and the cross entropy loss) by $g+\sigma_1, \ldots g+\sigma_D$. Here, again $g$ is the overall training set gradient and $\sigma_i$ is the noise (i.e.\ $\sum_{i=1}^{D} \sigma_i = 0$) of individual images (rather than mini-batches). Then the Empirical Fisher is given by 
\[
\text{EF} = \frac{1}{D}\sum_{i=1}^D (g+\sigma_i)^2 = g^2 + \EE[\sigma_k^2],
\] 
where $k$ is uniformly drawn from $\{1,\ldots,D\}$

\textbf{Second Moment Estimate of Fisher.} We want to compare EF to evaluating the squared gradient of a minibatch. Let $i_1, \ldots, i_b$ denote uniformly random, independent indices from $\{1,\ldots, D\}$, so that $X_{i_1}, \ldots, X_{i_b}$ is a random minibatch of size $b$ drawn with replacement. Let $g+\sigma$ be the gradient on this mini-batch. We then have, taking expectations over the random indices, 
\begin{eqnarray*}
	\EE[(g+\sigma)^2] &=& \EE\left[ \frac{1}{b^2}\sum_{r,s=1}^b (g+\sigma_{i_r})(g+\sigma_{i_s})\right] \\
	&=& \frac{b(b-1)}{b^2} \EE[(g+\sigma_{i_1})(g+\sigma_{i_2})]  +\frac{b}{b^2}\EE[(g + \sigma_{i_1})^2]\\
	&=& \frac{b-1}{b} g^2 + \frac{1}{b} \text{EF}\\
	&\approx& \frac{1}{b} \text{EF}
\end{eqnarray*}
The last approximation is biased, but it is still a decent approximation as long as $\EE[\sigma_k^2] \gg bg^2$. This explains why the second moment estimate of the Fisher gets worse for large batch sizes. Note that here $\sigma_k$ refers to the noise of the gradient with batch size 1 (whereas $\sigma$ is the noise of a mini-batch, which is $b$ times smaller).\\
For an analysis assuming that the mini-batch is drawn without replacement, see e.g.\ \citet{khan2018fast}. Note that for a minibatch size of 2048 and a trainingset size of 60000, the difference between drawing with or without replacement is small, as on average there are only 35  duplicates in a batch with replacement, i.e.\ less than $2\%$ of the batch. 

\subsection{Improved Estimate of Fisher for SOS.} 
We use the same notation as above, in particular $\sigma_k$ refers to the gradient noise of a single randomly sampled image (not an entire minibatch). Let us denote by $g+\sigma$ and $g+\sigma'$ the gradient estimates obtained from two independent minibatches of size $b$ (each sampled with replacement as above). Then for any $\alpha\in\mathbb{R}$, following the same calculations as above and slightly rearranging gives
\begin{eqnarray*}
	\EE\left[\bigl((g+\sigma) - \alpha(g+\sigma')\bigr)^2\right] &=& (1-\alpha)^2 g^2 + \frac{1+\alpha^2}{b}\EE[\sigma_k^2],
\end{eqnarray*}
where we used that each minibatch element is drawn independently and that $\EE[\sigma_k]=0$.
Now, if $(1-\alpha)^2 = \frac{1+\alpha^2}{b}$, then the above expression is proportional to the Empirical Fisher $\text{EF} = g^2 + \EE[\sigma_k^2]$. Since we rescale the regularisation loss with a hyperparameter, proportionality is all we need to get a strictly better (i.e.\ unbiased) approximation of the Empirical Fisher. 

The above condition for $\alpha$ and $b$ is satisfied when $\alpha = \frac{b+\sqrt{2b-1}}{b-1}$. In practice, we used $\alpha = 0$ for the experiments with batchsize 256 and $\alpha = 1$ for the experiments with batchsize 2048. Note that $\alpha=1$ is always a good approximation when $\EE[\sigma_k^2] \gg g^2$ (which we've seen to be the case). In our implementation SOS required calculating gradients on two minibatches, thus doubling the number of forward and backward passes during training. However, one could simply split the existing batch in two halves to keep the number of forward and backward passes constant. For large batch experiments we found introducing $\alpha$ necessary to obtain as good performance as with smaller batch sizes. This also supports our claim that SI's and SOS's importance measures rely on second order information contained in the Fisher.

\section{SI as Deacying Average}\label{sec:SI_SOS-better}
We concluded in the main paper that 

\begin{align}\label{eq:SI-SOS-good}
\tilde{\omega}(\SI) 
&\stackrel{\textbf{(A1)}}{\approx}
(1-\beta_1) \sum_{t\le T} \frac{\eta_t (g_t+\sigma_t)^2}{\sqrt{v_t}} \nonumber\\
&= \frac{1-\beta_1}{\sqrt{v_T}} \sum_{t\le T} \eta_t\sqrt{\frac{v_T}{v_t}} (g_t+\sigma_t)^2 \\
& \appropto \frac{1}{\sqrt{v_T}} v_T\nonumber
\end{align}
arguing that $\sqrt{v_T/v_t}$ is increasing (i.e.\ $v_t$ is decaying).

Upon first glance, it may seem that we could simply argue that $1/\sqrt{v_t}$ is increasing to conclude that 
\begin{align}
\tilde{\omega}(\SI) \label{eq:SI-SOS-bad}
&\stackrel{\textbf{(A1)}}{\approx}
(1-\beta_1) \sum_{t\le T} \eta_t\frac{1}{\sqrt{v_t}} (g_t+\sigma_t)^2  
 \appropto v_T
\end{align}
However, note that in this case the constant of proportionality hidden in this notation depends on the gradient magnitude - and thus it will be different for parameters with different gradient magnitudes, so that the overall approximation is bad.  To make this concrete, consider \eqref{eq:SI-SOS-bad} and multiply the stochastic gradients by 2. This will increase $(g+
\sigma_t)^2$ by a factor of 4, and $\sqrt{v_t}$ by a factor of 2, so the overall constant of proportionality will change by a factor of 2. 

In contrast, in equation \eqref{eq:SI-SOS-good} the constant of proportionality will be unaffected, showing that this first is independent of gradient magnitude and thus more likely to be accurate.
	\section{Exact Computation of Bias of SI with Adam}\label{sec:calc_bias}
We claimed that the difference between SI and SIU (green and blue line) seen in Figure \ref{figure:SI} (and also in Figures \ref{figure:summed_mnist_all}, \ref{figure:summed_cifar_all}) is due to the term $(1-\beta_1)\sigma_t^2$. To see this, recall that for SI, we approximate  $\frac{\partial L(t)}{\partial \theta}$ by $g_t+\sigma_t$, which is the same gradient estimate given to Adam. So we get
\begin{align*}
\textbf{SI:}&\qquad \frac{\partial L(t)}{\partial \theta} \Delta(t) = 
%(g+\sigma)\cdot \frac{(1-\beta_1)(g+\sigma) + \beta_1 m_{t-1}}{\sqrt{v_t} + \epsilon} \\
\frac{(1-\beta_1)(g_t+\sigma_t)^2}{{\sqrt{v_t} + \epsilon}} + \frac{\beta_1(g_t+\sigma_t)m_{t-1}}{{\sqrt{v_t} + \epsilon}}.\\
\intertext{For SIU, we use an independent mini-batch estimate $g_t+\sigma'_t$ for $\frac{\partial L(t)}{\partial \theta}$ and therefore obtain
} 
%\begin{eqnarray*}
\textbf{SIU:}&\qquad \frac{\partial L(t)}{\partial \theta} \Delta(t) =
%(g+\sigma)\cdot \frac{(1-\beta_1)(g+\sigma) + \beta_1 m_{t-1}}{\sqrt{v_t} + \epsilon} \\
\frac{(1-\beta_1)(g_t+\sigma'_t)(g_t+\sigma_t)}{{\sqrt{v_t} + \epsilon}} + \frac{\beta_1(g_t+\sigma'_t)m_{t-1}}{{\sqrt{v_t} + \epsilon}}.\\
\intertext{Taking the difference between these two and ignoring all terms which have expectation zero (note that $\EE[\sigma_t]=\EE[\sigma'_t] = 0$ and that $\sigma_t, \sigma'_t$ are independent of $m_{t-1}$ and $g_t$) gives
}
\textbf{SI}-\textbf{SIU:}&\qquad (1-\beta_1)\frac{\sigma_t^2}{\sqrt{v_t}+\epsilon}
\end{align*}
as claimed.

Note also that in expectation SIU equals $(1-\beta_1)\frac{g_t^2}{\sqrt{v_t}+\epsilon} + \beta_1\frac{g_t m_{t-1}}{\sqrt{v_t}+\epsilon}$ so that a large difference between SI and SIU really means $(1-\beta_1)\sigma_t^2 \gg (1-\beta_1)g_t^2 + \beta_1 m_{t-1}g_t \approx \beta_1 m_{t-1} g_t$. The last approximation here is valid because $\beta_1m_{t-1}\gg (1-\beta_1)g_t$ which holds since (1) $\beta_1 \gg (1-\beta_1)$ and (2) $\EE[|m_{t-1}|]\geq g_t$ since $\EE[|m_{t-1}|]\geq \EE[m_{t-1}] \approx \EE[|g_t|]$.

% and that $m_{t-1} \approx g_t$ (certainly they will be of a similar order of magni)

	\section{Experimental details}\label{sec:exp_detail}
%We tried to include all details necessary to reproduce our results here. If we forgot something, don't hesitate to send us an email. Code will also be available on github.

\subsection{Details of EWC and MAS related algorithms.}\label{sec:ewc_details}
For EWC we calculate the `real' Fisher Information as defined in the main article. For P-MNIST, we randomly sample 1000 training images (rather than iterating through the entire training set, which is prohibitively expensive). For Split CIFAR we use 500 random training images.

MAS, MAS-X and AF algorithms are also based on 1000 resp 500 random samples. When comparing these algorithms, we use the same set of samples for each algorithm.

\subsection{Benchmarks}
In \textbf{Permuted MNIST} \citep{goodfellow2013}  each task consists of predicting the label of a random (but fixed) pixel-permutation of MNIST. We use a total of 10 tasks. As noted, we use a domain incremental setting, i.e.\ a single output head shared by all tasks. 

In \textbf{Split CIFAR10/100} the network first has to classify CIFAR10 and then is successively given groups of 10 (consecutive) classes of CIFAR100. As in \citep{zenke2017}, we use 6 tasks in total. Preliminary experiments on the maximum of 11 tasks showed very little difference. We use a task-incremental setting, i.e.\ each task has its output head and task-identity is known during training and testing. 

\subsection{Pre-processing}
Following \citet{zenke2017}, we normalise all pixel values to be in the interval $[0,1]$ (i.e.\ we divide by 255) for MNIST and CIFAR datasets and use no data augmentation. We point out that pre-processing can affect performance, e.g.\ differences in pre-processing are part of the reason why results for SI on Permuted-MNIST in \citep{hsu2018} are considerably worse than reported in \citep{zenke2017} (the other two reasons being an unusually small learning rate of $10^{-4}$ for Adam in \citep{hsu2018} and a slightly smaller architecture).

\subsection{Architectures and Initialization}
For P-MNIST we use a fully connected net with ReLu activations and two hidden layers of 2000 units each. For Split CIFAR10/100 we use the default Keras CNN for CIFAR 10 with 4 convolutional layers and two fully connected layers, dropout \citep{srivastava2014dropout} and maxpooling, see Table \ref{table:arch}.
We use Glorot-uniform \citep{glorot2010understanding} initialization for both architectures.

\subsection{Optimization}
We use the Adam Optimizer \citep{kingma2014adam} with tensorflow \citep{abadi2016tensorflow} default settings ($lr=0.001, \beta_1=0.9, \beta_2 = 0.999, \epsilon = 10^{-8}$) and batchsize 256 for both tasks like \citet{zenke2017}. 
On Permuted-MNIST we train each task for 20 epochs, on Split CIFAR we train each task for 60 epochs, also following \citet{zenke2017}. 
We reset the optimizer variables (momentum and such) after each task. 

For the two experiments confirming our prediction regarding the effect of batchsize on SI, we made two changes on top of using a batch size 0f 2048: On P-MNIST we increased the learning rate by a factor of 8, since the ratio of batchsize and learning rate is thought to influence the flatness of minima (e.g.\ \cite{jastrzkebski2017three, mirzadeh2020understanding}). We did not tune or try other learning rates. On CIFAR, we tried the same learning rate adjustment but found that it led to divergence and we resorted back to the default 0.001. We observed that with a batch size of 2048, already on task 2, the network did not converge during 60 epochs (for tasks 2-6, one epoch corresponds to only 2 parameter updates with this batch size). We therefore increased the number of epochs to 600 on tasks 2-6, to match the number of parameter updates of Task 1 (CIFAR10).

\subsection{SI \& SOS details}\label{sec:SI_details}
Recall that we applied the operation $\max(0,\cdot)$ (i.e.\ a ReLU activation) to the importance measure of each individual task (\eqref{eq:rescale} of main paper), before adding it to the overall importance. In the original SI implementation, this seems to be replaced by applying the same operation to the overall importances (after adding potentially negative values from a given task). No description of either of these operations is given in the SI paper. In light of our findings, our version seems more justified.

Somewhat naturally, the gradient $\frac{\partial L}{\partial \theta}$ usually refers to the cross-entropy loss of the current task and not the total loss including regularisation. For CIFAR we evaluated this gradient without dropout (but the parameter update was of course evaluated with dropout).\footnote{We did not check the original SI code for what's done there and the paper does not describe which version is used.}

For SOS, similarly to SI, we used the gradient evaluated without dropout on the cross-entropy loss of the current task for our importance measure $\sqrt{v_t}$.

When evaluated on benchmarks, SI, SIU, SIB all rescaled according to \eqref{eq:rescale} to make the comparison as fair as possible. Scatter plots and correlations in the main text also use this rescaling, as this is the quantity used by the algorithms. In Appendix \ref{moreisbetter} we also include results before rescaling.

\begin{table}[h!]
	\caption{ CIFAR 10/100 architecture. Following \citet{zenke2017} we use the keras default architecture for CIFAR 10.  
		Below, `Filt.' refers to the number of filters of a convolutional layer, or respectively the number of neurons in a fully connected layer. `Drop.' refers to the dropout rate. `Non-Lin.' refers to the type of non-linearity used. \\
		Table reproduced and slightly adapted from \citep{zenke2017}.}
	\label{table:arch}
	\vskip 0.05in
	\begin{center}
		\begin{small}
			\begin{tabular}{r|ccccc}
				\toprule
				Layer & Kernel & Stride & Filt. & Drop. & Non-Lin. \\
				\midrule
				3x32x32 input &&&&& \\
				Convolution & 3x3 & 1x1 &32&& ReLU\\
				Convolution & 3x3  & 1x1 &32&& ReLU\\
				MaxPool & 2x2 & 2x2 &&0.25&\\
				Convolution & 3x3  & 1x1 &64&& ReLU\\
				Convolution & 3x3  & 1x1 &64&& ReLU\\
				MaxPool & 2x2 & 2x2 &&0.25&\\
				FC &  &  &512&0.5&ReLU\\
				\midrule
				Task 1: FC &  &  &10&&softmax\\
				...:FC &  &  &10&&softmax\\
				Task 6: FC &  &  &10&&softmax\\
				\bottomrule
			\end{tabular}
		\end{small}
	\end{center}
	\vskip -0.1in
\end{table}

\iffalse
\section{Ablation for Normalisation Term}\label{sec:ablation}
Recall that in Table \ref{table:results}, Exp.\ No.\ 3, there is a gap in performance between SI, OnAF (both 74.4\% accuracy) and AF (73.4\% accuracy). Part of this difference may stem from the fact that the first two algorithms are evaluated online along the training trajectory, while AF is only evaulated at the end of training. Additionally, there may be a difference in performance stemming from the normalisation of OnAF and SI, c.f. equation \eqref{eq:rescale} of the main paper, which is not applied to AF. To assess this difference, we tested a version of OnAF which is not normalised as in \eqref{eq:rescale}. It achieved an average accuracy of 
73,8$\pm$ 0.8
on Split CIFAR.

Furthermore, we point out that we consistently observed slight improvements of SI, OnAF, SIB on P-MNIST and CIFAR when the normalisation was included.
\fi

\subsection{Hyperparameters}\label{sec:HP}
For all methods and benchmarks, we performed grid searches for the hyperparameter $c$  and over the choice whether or not to re-initialise model variables after each task.

The grid of $c$ included values $a\cdot 10^i$ where $a\in\{1,2,5\}$ and $i$ was chosen in a suitable range (if a hyperparameter close to a boundary of our grid performed best, we extended the grid). 

For CIFAR, we measured validation set performance based on
at least three repetitions for good hyperparameters. We then picked the best HP and ran 10 repetitions on the test set. For MNIST, we measured HP-quality on the test set based on at least 3 runs for good HPs.

Additionally, SI and consequently SIU, SIB have rescaled importance (c.f.\ equation \eqref{eq:rescale} from main paper). The damping term $\xi$ in this rescaling was set to $0.1$ for MNIST and to $0.001$ for CIFAR following \citet{zenke2017} without further HP search. 

All results shown are based on the same hyperparameters obtained -- individually for each method -- as described above. They can be found in Table \ref{table:HP}. We note that the difference between the HPs for MAS and MASX might seem to contradict our claims that the two measures are almost identical (they should require the same $c$ in this case), but this is most likely due to similar performance for different HPs and random fluctuations. For example on CIFAR, MAS had almost identical validation performance for $c=200$ (best for MAS) and $c=1000$ (best for MASX) ($74.2\pm0.7$ vs $74.1\pm 0.5$).

Also for the other methods, we observed that usually there were two or more HP-configurations which performed very similarly. The precise `best' values as found by the HP search and reported in Table \ref{table:HP} are therefore subject to random fluctuations in the grid search.
\begin{table}
	\caption{Hyperparameter values for our experiments. See also maintext.}
	\label{table:HP}
	\vskip -0.1in
	\begin{center}
		\begin{small}
			\begin{tabular}{r|cc|cc}
				\toprule
				\multirow{2}*{\textbf{Algorithm}} & \multicolumn{2}{c}{\textbf{MNIST}} & \multicolumn{2}{c}{\textbf{CIFAR}}\\
				&\textbf{c}& \textbf{re-init}&\textbf{c}&\textbf{re-init} \\
				\midrule
				SI & 0.2 & $\surd$ & 5.0 & $\surd$ \\
				SI(2048) & 1.0 & $\surd$ & 20.0 & $\surd$ \\
				SIU & 2.0 & $\surd$ & 2.0 & $\times$ \\
				SIB & 0.5 & $\surd$ & 5.0 & $\surd$ \\
				SOS & 1.0 & $\surd$ & 500 & $\surd$ \\
				SOS(2048) & 10 & $\surd$ & 2000 & $\surd$ \\
				MAS & 500 & $\times$ & 200 & $\surd$ \\
				AF & 200 & $\times$ & 1e3 & $\surd$ \\
				$\sqrt{\text{Fisher}}$ & 2 & $\times$ & 10 & $\surd$ \\				
				EWC & 1e3 & $\times$ & 5e4 & $\surd$ \\
				MAS2 (logits) &  0.001& $\surd$ & 0.1 & $\surd$ \\
				\bottomrule
			\end{tabular}
		\end{small}
	\end{center}
	\vskip -0.2in
\end{table}

\subsection{Scatter Plot Data Collection }\label{sec:intensity}
The correlation and scatter plots are based on a run of SI, where we in parallel to evaluating the SI importance measure also evaluated all other importance measures. Correlations in Figures \ref{figure:MAS} and \ref{figure:SI} show data based on 5 repetitions (reporting mean and std-err as error bars) of this experiment and confirm that the correlations observed in the scatter plots are representative. 
%In addition, we gathered data following the precise setup of \citet{zenke2017}, i.e.\ without reinitialising model variables after each task, and found even stronger support for our heuristics in this setting, see Section \ref{sec:effect_reg_SI}. 

%\paragraph{Intensity Plots} (as used for some figures in the appendix) Show same kind of data as scatter plots, but each weight is binned into one of 80x50 equispaced bins. The number of weights per bin is divided by the maximum number of weights per bin in that column and the resulting value is shown on a gray scale. Normalisation per column was performed since for `global normalisation' (divide by total number of weights) only weights in the bottom left are visible (as most weights have small importances); the number of bins was chosen to match aspect ratio of plots.

\subsection{Computing Infrastructure}
We ran experiments on one NVIDIA GeForce GTX 980 and on one NVIDIA GeForce GTX 1080 Ti GPU. We used tensorflow 1.4 for all experiments.

\section{The Fisher and its square root for posterior variance}\label{sec:sqrtF}
Suppose that we know a locally optimal set of parameters $\theta_{opt}$. Then we can use a Laplace approximation around $\theta_{opt}$ to find the likelihood of our parameters $\theta \sim \mathcal{N}(\theta_{opt}, K)$, where $K$ is the (diagonal) inverse Hessian, or Fisher Information. 

It is customary to use the final parameter point $\theta_{final}$(after training) as approximation of $\theta_{opt}$, which also leads to the EWC algorithm. However, if we have uncertainty estimates for $\theta_{opt}$ we can (and should?) incorporate these into the model.
Assuming $\theta_{opt} \sim \mathcal{N}(\theta_{final}, R)$,  with ${p(\theta) = \int p(\theta \mid \theta_{opt})  \cdot p(\theta_{opt})d\theta_{opt}}$, we get 

\[
\theta \sim \mathcal{N}(\theta_{final}, K+R)
\]
where we assumed that the inverse Hessian $K$ is constant across values of $\theta_{opt}$ (which is an assumption already made in \citep{huszar2018} providing a Bayesian justification for EWC).

Recently, \citet{aitchison2018bayesian} presented a Kalman filter model for optimization, whose uncertainty estimates for $\theta_{opt}$, recover the Adam Optimizer \citep{kingma2014adam} and we now show how to plug in this model in the above equations.

 For sake of simplicity we take $\sqrt{\eta^2 / \langle g^2 \rangle}$ as posterior variance of $\theta_{opt}$, where $\langle g^2 \rangle = v_T$ is the exponentially decaying average of the second moment of the stochastic gradients. In the framework of \cite{aitchison2018bayesian} this corresponds to a uniform prior (which, judging from the absence of weight decay, is used by EWC, SI). \footnote{Note that in addition to a uniform prior, this also relies on the assumption that gradients are not too large, by which we mean that $\sqrt{v_T}$ is not much bigger than $\sim 1000$, since otherwise the approximation in equation (47) of \cite{aitchison2018bayesian} becomes inaccurate due to large values of $\lambda_{post}$.}

Now, using this framework, we plug in suitable values for $K, R$. 
Let's denote the mini-batch size by $b$ and the exponentially decaying average of the gradient's second moment by $\langle g^2 \rangle$. Then the Hessian is approximately $b\cdot\langle g^2\rangle$, giving $K = 1/(b\cdot \langle g^2\rangle)$. As discussed above, the variance $R$ of $\theta_{opt}$ is given by $\sqrt{\eta^2 / \langle g^2 \rangle}$, where $\eta=0.001$. 
It may seem that we should also multiply $g^2$ by $b$ since $g^2$ serves as an approximation for the Hessian here, too. However, the batch size was already compensated for by the choice of $\eta$ as long as we use a `standard' batch size, let's say $b\sim 100$ (see \citet{aitchison2018bayesian} and below).

Altogether, we get a posterior variance of the likelihood of $\theta$ of
\[
K+R = \frac{1}{b\langle g^2 \rangle} + \frac{\eta}{\sqrt{\langle g^2\rangle}}  
\]
Using the inverse variance as an importance measure for continual learning, we get 
\[
\frac{1}{K+R} = \frac{b \langle g^2 \rangle}{1 + b\eta \sqrt{\langle g^2 \rangle}}.
\]

For $b\eta\sqrt{\langle g^2\rangle}>1$ ( or $\sqrt{\langle g^2\rangle} \sim 10$ and $b\sim100$), this importance (as well as the posterior variance) is approximated better by the square-root $\sqrt{\langle g^2 \rangle}$ than by $\langle g^2 \rangle$ itself. 
As already pointed out in the main text, this is not quite the regime we operate in on Permuted-MNIST and Split CIFAR with the given architectures.

\subsection{Dependence on Batch Size}
This subsection assumes familiarity with \citep{aitchison2018bayesian} and specifically, the argument used there to choose $\eta$.

To make the dependence on the minibatch size clearer, we can use $b\langle g^2 \rangle$ everywhere as approximation of the Hessian, leading to $\sqrt{\eta'^2 / (b\langle g^2 \rangle)}$ as posterior variance of $\theta_{opt}$. Assuming $b\sim100$ we should then set $\eta'=0.01$ instead. This results in

\[
K+R = \frac{1}{b\langle g^2 \rangle} + \frac{\eta'}{\sqrt{b\langle g^2\rangle}}  
\]
and 
\[
\frac{1}{K+R} = \frac{b \langle g^2 \rangle}{1 + \eta' \sqrt{\langle bg^2 \rangle}}.
\]

		\section{MAS based on logits}\label{sec:MAS_logits}
	As pointed out in Section \ref{sec:desc}, there are two versions of MAS. Here, we describe results related to the version based on logits. We note that this version has the potentially undesirable feature of not being invariant to reparametrisations: If, for example, we add a constant $c$ to all the logits, this does not change predictions (or the training process), but it does change the logit-based MAS importance.

	We found identical performance as for the both versions of MAS (of course tuning HPs individually): $97.2\pm 0.1$ on P-MNIST and $73.9 \pm 0.2$ on CIFAR for logit based MAS, as compared to $97.3 \pm 0.1$ and $73.7 \pm 0.2$ for the version based on the probability-distribution output. For both benchmarks, the difference based on 10 repetitions was not statistically significant ($p>0.4$ for both benchmarks, t-test).
	
\begin{figure}[h]
		%\vskip -40pt
		\begin{center}
			\begin{subfigure}{0.3\textwidth}
				%\vskip 0.1in
				\centerline{\includegraphics[width=\textwidth]{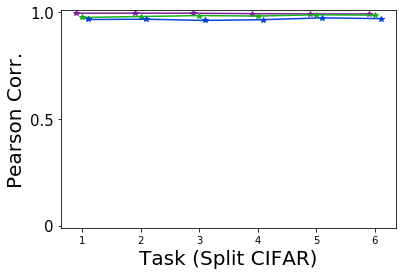}}
			\end{subfigure}
			\hskip 6pt
			\begin{subfigure}{0.3\textwidth}
				%\vskip 0.1in
				\centerline{\includegraphics[width=\textwidth]{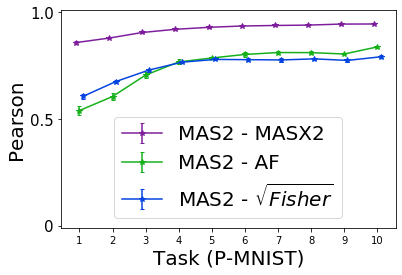}}
			\end{subfigure}
			\hskip 6pt
			\begin{subfigure}{0.33\textwidth}
				\vskip -0.15in
				\centerline{\includegraphics[width=\textwidth]{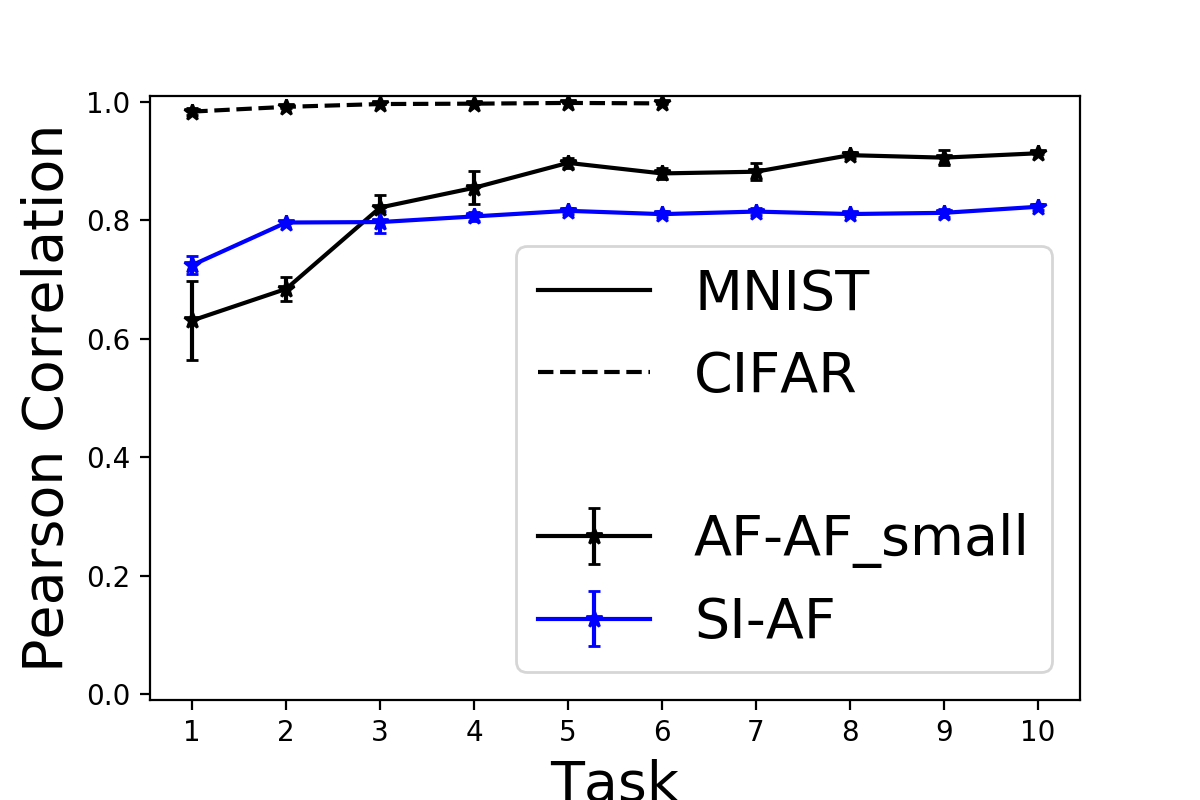}}
			\end{subfigure}		
		\end{center}
		\caption{
			\textbf{Empirical Relation between MAS (logits) and Square Root Fisher.} \\
			\textbf{Left \& Mid:} Same as Figure \ref{figure:MAS}, left, but using MAS based on logits rather than output distribution.\\
			\textbf{Right:} Comparing AF based on different sample sizes. 1000 vs 30000 on MNIST and 500 vs 5000 on CIFAR. Comparing SI, which relies on gradients of all 60000 images, to AF using 30000 images. 
		}
		\label{figure:MAS_logits}
	\end{figure}

	Next, we investigated the relation between logit based MAS and the square root of the Fisher. On CIFAR we found correlations very close to 1 for all pairs and on MNIST correlations were slightly weaker, see Figure \ref{figure:MAS_logits}. 
	Note that on MNIST  the correlations between e.g.\ MAS and AF are similar to the correlation between AF and SI, Figure \ref{figure:MAS_logits} (right). 
	We also show how AF depends on sample size, comparing 1000 (used by MAS) to 30000 samples on MNIST and 500 to 5000 on CIFAR, see  Figure \ref{figure:MAS_logits} (right). 
	
	%We also mention that MAS based on logits bares some similarity with the Generalised Gauss Newton matrix.
		
	\section{Gradient Noise}\label{sec:noise}

Here, we quantitatively assess the noise magnitude outside the continual learning context. 
Recall that Figure \ref{figure:SI} (left) from the main paper, as well as Figures \ref{figure:summed_mnist_all} and \ref{figure:summed_cifar_all} already show that the noise dominates the SI importance measure, which indicates that the noise is considerably larger than the gradient itself. 

\begin{figure}[h]
	\vskip -8pt
	\begin{center}
		\begin{subfigure}{0.9\textwidth}
			\centerline{\includegraphics[width=\textwidth]{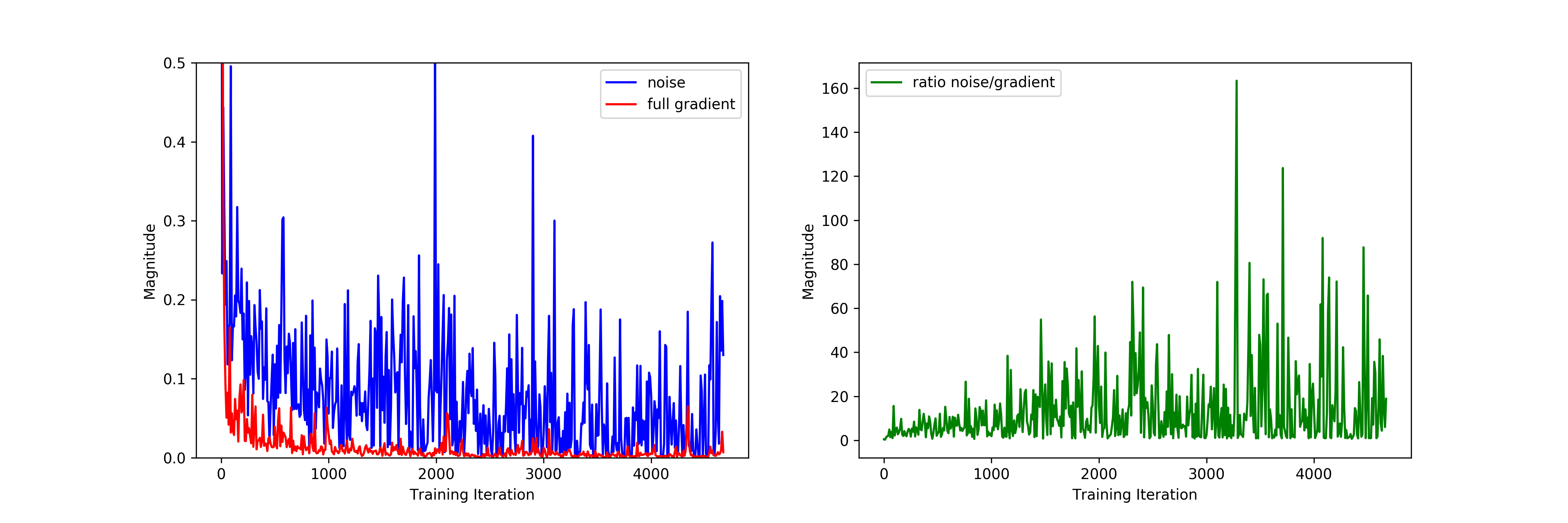}}
		\end{subfigure}
		\\
		\begin{subfigure}{0.5\textwidth}
			\centerline{\includegraphics[width=\textwidth]{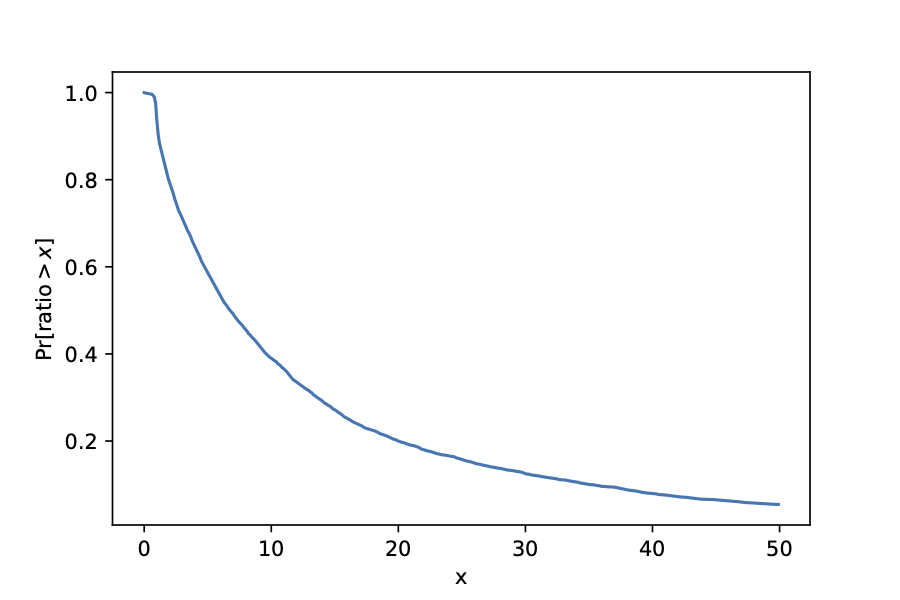}}
		\end{subfigure}
	\end{center}
	\vskip -0.2in
	\caption{\textbf{Top:} Gradient noise, measured as squared $\ell_2$ distance between full training set gradient and the stochastic mini-batch gradient with a batch size of 256. `Full gradient' magnitude is also measured as squared $\ell_2$ norm.\\
		Data obtained by training a ReLu network with 2 hidden layers of 2000 hidden units for 20 epochs with default Adam settings. Only every 20-th datapoint shown for better visualisation.\\
		\textbf{Bottom:} Same data. $y$-value shows fraction of training iterations in  which the ratio between mini-batch noise and full training set gradient was at least $x$-value. In particular the batch-size was 256. `Ratio' refers to the ratio of squared $\ell_2$ norms of the respective values.}
	\label{figure:noise}
	\vskip -0.1in
\end{figure}

To obtain an assessment independent of the SI continual learning importance measure, we trained our network on MNIST as described before, i.e.\ a ReLu network with 2 hidden layers of 2000 units each, trained for 20 epochs with batch size 256 and default Adam settings. At each training iteration, on top of calculating the stochastic mini-batch gradient used for optimization, we also computed the full gradient on the entire training set and computed the noise --  which refers to the squared $\ell_2$ distance between the stochastic mini-batch gradient and the full gradient -- as well as the ratio between noise and gradient, measured as the ratio of squared $\ell_2$ norms. The results are shown in Figure \ref{figure:noise} (top). In addition, we computed the fraction of iterations in which the ratio between noise and squared gradient norm is above a certain threshold, see Figure \ref{figure:noise} (bottom).

	\section{Relative Performance of Regularisation Methods}\label{sec:reg}
Several papers \citep{van2019, hsu2018, farquhar2018} played an important role in recognising that different continual learning algorithms had been evaluated in settings with varying difficulties. These papers also greatly clarified these settings, making an important contribution for future work. 

Here, we want to point out that some of the results (and thus conclusions) of the experiments in \citep{hsu2018} (these were the only ones we investigated more closely) are possibly specific to settings and hyperparameters used there. 

For example, it is found there that EWC, MAS, SI perform worse than an importance measure which assigns the same importance to all parameters (called `L2 regularisation'). This is in contrast to results reported in the EWC paper \citep{kirkpatrick2017}, who found the L2 baseline to perform worse than EWC. We also attempted to reproduce the L2 results and tried two versions of this approach, one with constant importance for all tasks and one with importance increasing linearly with the number of tasks. The latter approach worked better, but still considerably underperformed EWC, MAS, SI on Permuted-MNIST (88\% vs 97\%). 
We didn't run these experiments on CIFAR. An alternative importance measure based on the weights' magnitudes after training was better, but still could not compete with SI, EWC, MAS (<95\% vs >97\%) showing that these importance measure are more use- and meaningful than naive baselines.

Additionally, in other work %\citep{stochasticsynapses} 
%[no citation for anonymity]
we found that for domain incremental settings (again, these were the only ones we investigated), results of baselines EWC and SI, but also of fine-tune (a baseline which takes no measures to prevent catastrophic forgetting) can be improved, often by more than 10\% compared to previously mentioned reports \citep{hsu2018, van2019}. This also implies that the difference between well set-up regularisation approaches and for example repaly-methods is not nearly as big as previously thought. The differences between these results are explained by different factors (we explicitly tested each factor): hyperparameters (for example using Adam with the standard learning rate 0.001 improves performance over the learning rate 0.0001 used previously), initialisation (Glorot Uniform Initialisation works considerably better for SI than pytorch standard initialisation for linear layers, which reveals unexpected defaults upon close investigation), training duration (regularisation approaches usually notably benefit from longer training, this is reported in \cite{swaroop2019}, and we also found that SI's performance on P-MNIST can be improved to 98\% by training for 200 rather than 20 epochs; \cite{hsu2018} use very short training times), data-preprocessing (depending on the setting different normalisations have different, non-negligible impacts).

	\section{More Experiments and Plots}\label{moreisbetter}
Here, we include additional plots and experiments.

\subsection{SI with SGD}
We tested SGD+momentum on Permuted MNIST for SI. We found that SI has similar performance as before and again is better SIU (which is also similar to before). However, training was more unstable and occasionally diverged for both SI an SIU, probably due to the ill-conditioned optimization caused by the regularisation loss, so that we did not run SGD experiments on CIFAR.

\subsection{Bias of SI on All tasks}
\begin{figure}[h]
	%\vskip -0.1in
	\begin{center}
		\centerline{\includegraphics[width=1.2\textwidth]{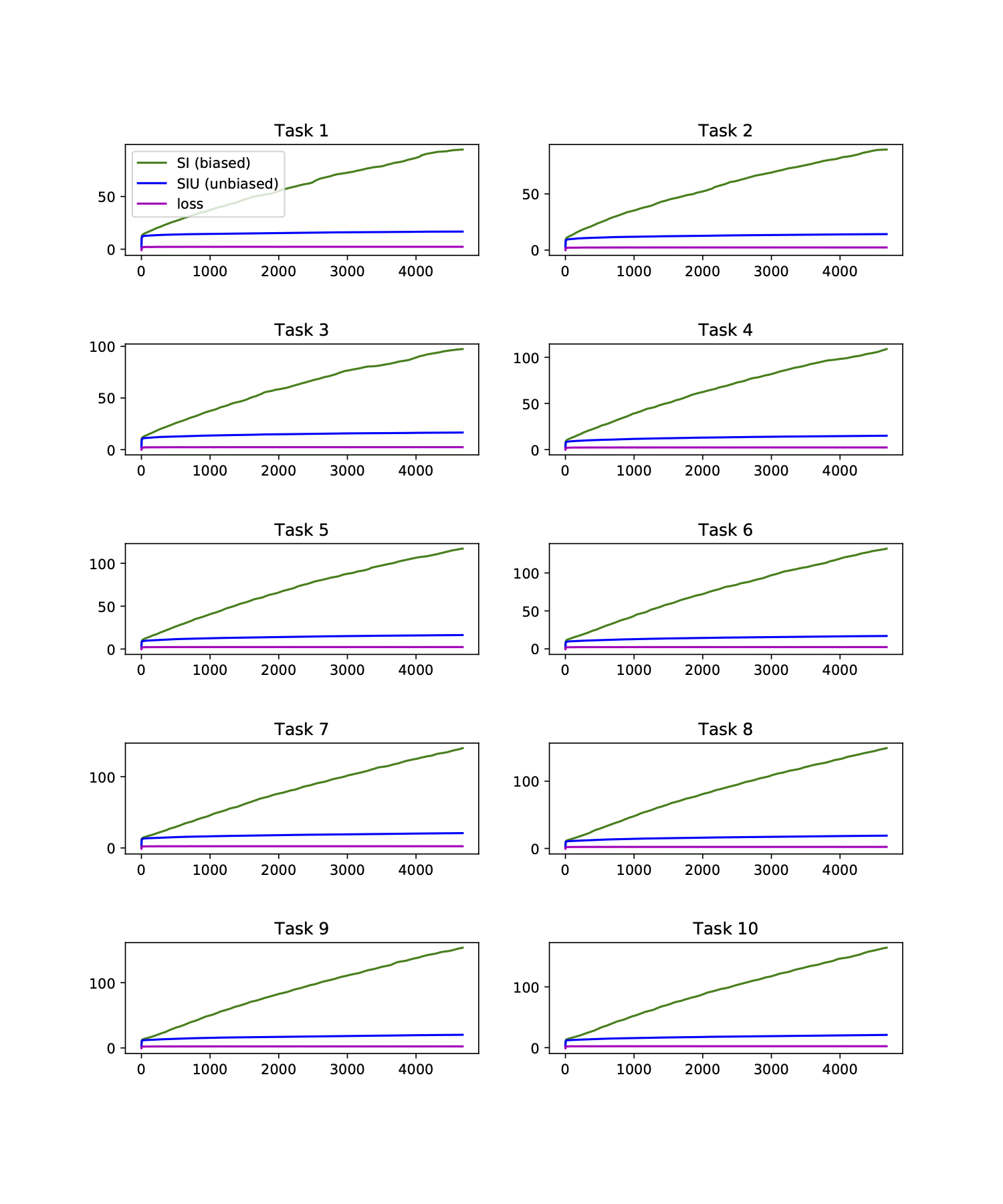}}
	\end{center}
	%\vskip -1.0in
	\caption{Summed importances for all P-MNIST tasks for SI and its unbiased version. Analogous to Figure \ref{figure:SI} (left) from main paper. Note and excuse changed color coding of SI, SIU.}
	\label{figure:summed_mnist_all}
	%\vskip -0.4in
\end{figure}

\begin{figure}[t!]
	%\vskip -1.1in
	\begin{center}
		\centerline{\includegraphics[width=1.2\textwidth]{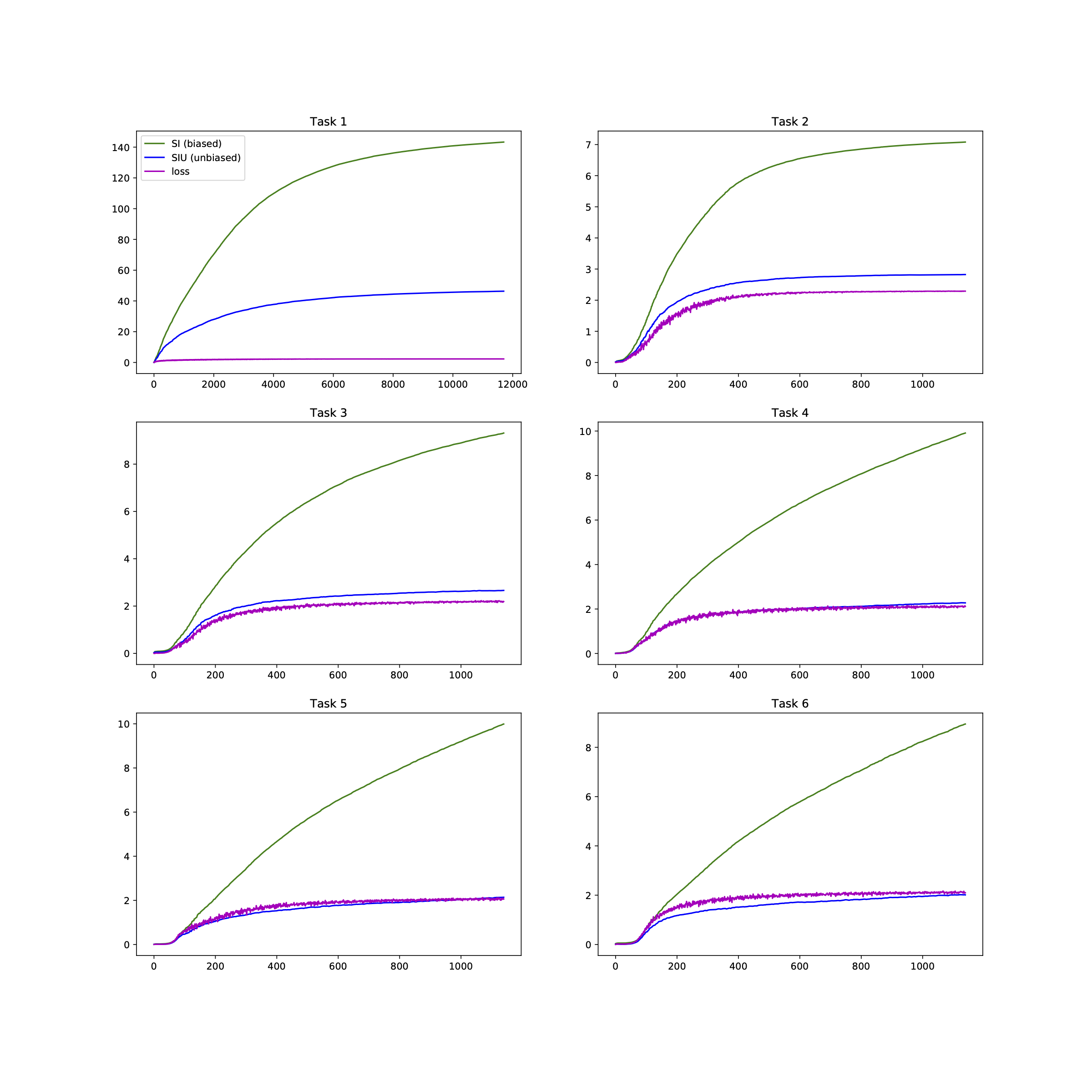}}
	\end{center}
	%\vskip -1.0in
	\caption{Summed importances for all Split CIFAR 10/100 tasks for SI and its unbiased version. Analogous to Figure \ref{figure:SI} (left) from main paper. Note and excuse changed color coding of SI, SIU.}
	\label{figure:summed_cifar_all}
	%\vskip -0.1in
\end{figure}

Here, we report the magnitude of the bias of SI on all tasks, analogously to Figure \ref{figure:SI}, left panel.
The observation that most of the SI importance is due to its bias is consistent across datasets and tasks, see Figures \ref{figure:summed_mnist_all} and \ref{figure:summed_cifar_all}. Intriguingly, on CIFAR we find that the unbiased approximation of SI slightly underestimates the decrease in loss in the last task, suggesting that strong regularisation pushes the parameters in places, where the cross-entropy of the current task has negative curvature.

We did not conduct the analogue of experiment corresponding to the controls in Figure \ref{figure:SI}, right panel, on MNIST since there the influence of regularisation was already weak.

%\subsection{Additional Scatter Plots and Intensity Plots}
%We show additional scatter and intensity plots in Figure \ref{fig:scatter_app}. Note that for both architectures there are a few million weights. The scatter plots are overcrowded and show the whole range of dependencies between two measure that can possibly occur. Intensity plots show the dependencies that the majority of weights adheres to.

\subsection{Comparison of SI, SIU, SIB to SOS before rescaling}
The comparisons between SI, SIU, SIB and SOS in the main paper were obtained before rescaling in  \eqref{eq:rescale}. 

\begin{figure}[h]
\begin{center}
	\begin{subfigure}{0.3\textwidth}
		%\vskip 0.1in
		\centerline{\includegraphics[width=\textwidth]{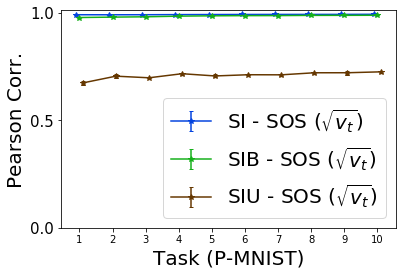}}
	\end{subfigure}
	\hskip 12pt
	\begin{subfigure}{0.3\textwidth}
		%\vskip 0.1in
		\centerline{\includegraphics[width=\textwidth]{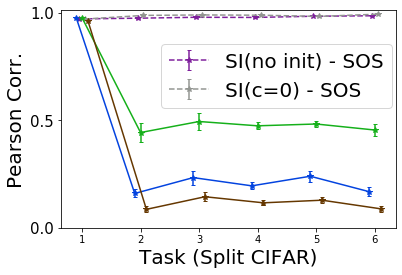}}
	\end{subfigure}
\end{center}
\caption{Analogous to Figure \ref{figure:SI} from main paper, comparing SI,SIU, SIB to SOS, but before applying the rescaling in \eqref{eq:rescale} to SI, SIU, SIB (but not SOS).
Note that rescaling on MNIST has close to no effect, which is due to almost all importances being non-negative and the denominator in \eqref{eq:rescale} being almost equal to the damping term $\xi=0.1$. On CIFAR, rescaling increases correlations on Tasks 2-5, mostly due to the $\max(\cdot,0)$ operation. On Task 1, correlation to SOS is decreased by rescaling, as expected, due to division in \eqref{eq:rescale}. There are no negative importances after first task, due to absence of the regularisation term.\\
Note that it may seem surprising that on the first task of Split CIFAR (i.e.\ CIFAR10) all of SIU, SIB, SI have similarly high correlations to SOS in this figure as well as Figure \ref{figure:SI}. 
However, similarity between SIU before rescaling and SOS in this situation is explained by the weights with largest importance:
If we remove the 5\% of weights (results similar for 1\%) which have highest SIU importance, correlation between SOS, SIU drops to $0.5$, but for SI resp.\ SIB the same procedure yields $0.9$. This indicates that for the first task of CIFAR there is a small fraction of weights with large importances, which are dominated by the gradient rather than the noise. The remainder of weights is in accordance with our previous observations and dominated by noise, recall also Figure \ref{figure:summed_cifar_all}.
}
\label{figure:correlations_SI_divide}
\end{figure}

Results before rescaling are shown and discussed in Figure \ref{figure:correlations_SI_divide}, also confirming that correlation between SOS and SI is due to the bias of SI.

\end{document}